\documentclass[letterpaper]{article}
\usepackage{aaai2027}
\nocopyright
\usepackage[hyphens]{url}
\usepackage{graphicx}
\usepackage{natbib}
\usepackage{caption}
\usepackage{booktabs}
\usepackage{amsmath,amssymb}
\title{When the Judge Acts: Auditing VLM-Guided Image Selection\\on Culturally Situated Prompts}
\author{Huichan Seo}
\affiliations{Independent Researcher\\
gmlcks00513@gmail.com}
\begin{document}
\maketitle
\begin{links}
\link{Code and project page}{https://github.com/seochan99/JudgeActs}
\end{links}
\begin{abstract}
Vision-language models (VLMs) increasingly act as judges that pick the best of several generated images, so their choices decide what users see. Such judges are usually validated by score agreement with human ratings, not by the images they return. We audit VLM judges as decision-makers: on 300 culturally situated prompts, we compare the returned image with human ratings the judge never sees and with random choice from the same candidates, and repeat every decision with the candidates reordered. A 4B-parameter judge barely beats random and falls short of a CLIP similarity baseline. It picks the first image shown in 49\% of calls (chance: 28\%), and reordering changes its choice on 60\% of prompts. For this judge, agreement across orders is informative: decisions that survive reordering are much better than random, whereas agreement with a weaker second judge keeps the wrong ones. An 8B judge shows almost no position bias and outperforms CLIP, yet for it the same filter mostly discards good decisions. Agreement helps only when it targets the judge's failure mode, so filters must be re-audited whenever the judge changes. The 4B judge's slight rise in stereotype ratings is no longer detectable after aggregating across orders or with the larger judge.

\end{abstract}
% Both frozen main runs are complete; this is not an upload confirmation.

\section{Introduction}
A vision-language model (VLM) that selects one of several generated images does more than report a quality score. Its choice determines which candidate a user would receive. The practical question is whether that choice improves the available outputs, not just whether the judge's scores correlate with human ratings.

For culturally situated requests, visual polish and recognizable symbols can be poor guides to prompt satisfaction. An image may include a requested object while misrepresenting the social context around it. A selector can favor that image even when another candidate better satisfies the request. Evaluating this choice requires ratings that did not come from the selector itself.

CulturalFrames provides generated images and human ratings that distinguish missing explicit expectations, missing implicit expectations, and stereotypes. It also reports weak agreement between automatic metrics and human judgments \citep{culturalframes}. We use the public CulturalFrames release to study a different decision: choosing which existing image to return.

\begin{figure*}[t]
\centering
\includegraphics[width=\textwidth]{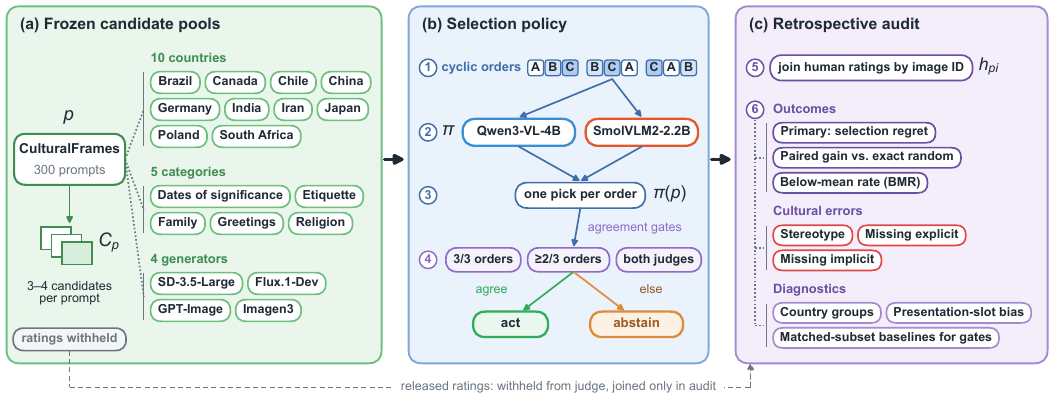}
\caption{Audit protocol. (a) Frozen candidate pools from CulturalFrames: culturally situated prompts with three or four images from four generators. (b) Selection: each judge sees only the prompt and the images, in three cyclic orders; agreement gates decide whether to act or abstain. (c) Retrospective audit: released human ratings, never shown to the judge, are joined by image identifier after selection and compared with exact random selection on the same pools.}
\label{fig:framework}
\end{figure*}

For each prompt, an open-weight VLM sees three or four candidate images and selects one (Figure~\ref{fig:framework}), without generator names, country metadata, or ratings. We compare the selected image's released ratings with exact random choice and with the best candidate in the same fixed pool, so differences between policies arise from selection rather than generation.

\begin{figure}[!b]
\centering
\includegraphics[width=\columnwidth]{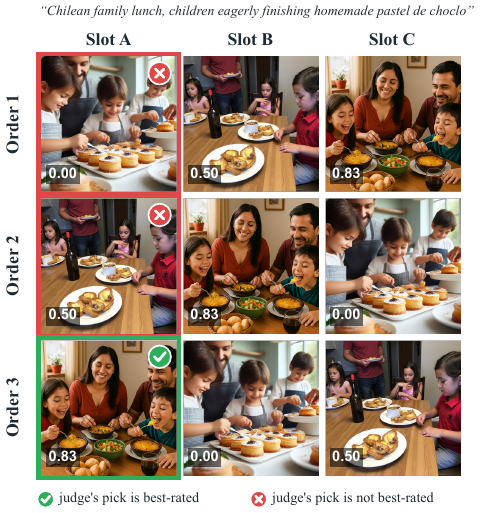}
\caption{Same request, same images, three orders. Each row shows the three candidates in one presentation order, with mean human prompt-alignment ratings. Qwen3-VL-4B picks the image in slot A every time (framed), so its returned image is rated 0.00, 0.50, or 0.83 depending only on order. Images: CulturalFrames \citep{culturalframes}.}
\label{fig:hero}
\end{figure}

The primary outcome is \emph{selection regret}, measured from the mean human prompt-alignment rating. It is the gap between the highest-rated available candidate and the selected image. Prompt alignment measures request satisfaction, not cultural accuracy alone. We therefore report stereotype and missing-expectation ratings separately. We also measure how often the selected score falls below random expected utility. We call this the \emph{below-mean rate} (BMR).

\begin{figure*}[t]
\centering
\includegraphics[width=\textwidth]{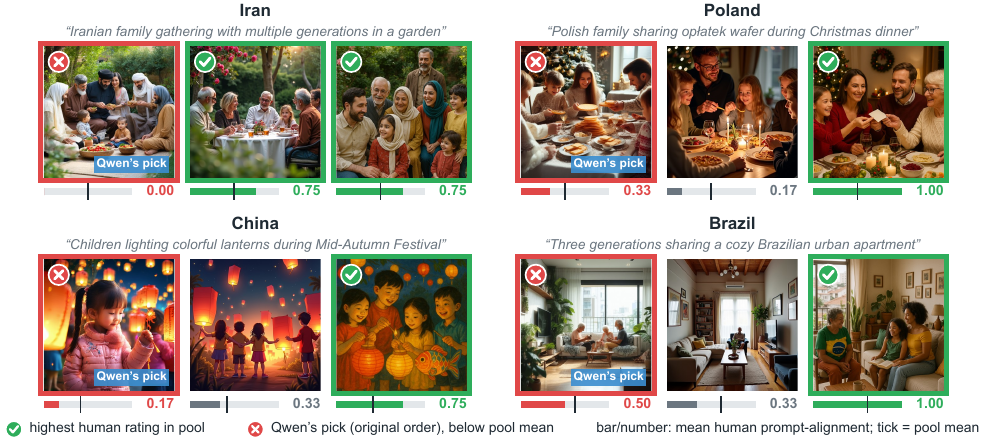}
\caption{VLM judges can return below-average images when a clearly better candidate exists. Each group is one CulturalFrames prompt \citep{culturalframes}; numbers are mean human prompt-alignment ratings. Red: Qwen3-VL-4B's original-order choice, below the pool mean. Green: highest-rated candidate. Prompts follow a fixed rule applied without viewing images (Appendix~\ref{app:gallery}); they illustrate the failure mode, not its frequency (Figure~\ref{fig:results}).}
\label{fig:teaser}
\end{figure*}

We make three contributions.
\begin{itemize}
\item \textbf{An audit of judges as decision-makers.} We score each returned image against held-out human ratings and against random choice from the same candidates, so every comparison is paired and free of the judge's own scores.
\item \textbf{How position bias changes what users receive.} Position bias in model judges is known; we measure its consequence when the judge acts. Both small judges favor early positions, and the same prompt can yield the worst or the best available image depending only on candidate order (Figures~\ref{fig:hero} and~\ref{fig:teaser}). Ablations show that the bias is not caused by our output format or label letters.
\item \textbf{An audit of agreement gates across judges.} For the 4B judge, keeping only decisions that are consistent across orders retains choices that beat random, whereas agreement with a weaker second judge does not. For a larger 8B judge with little position bias, the same gate mostly discards good decisions. A gate must therefore be audited on the prompts it rejects as well as on those it keeps, and re-audited when the judge changes.
\end{itemize}

\section{Related Work}
\paragraph{Cultural representation and evaluation.}
CulturalFrames supplies culturally situated prompts, multiple generated images, and ratings of missing expectations and stereotypes \citep{culturalframes}. Its analysis compares automatic metrics with human ratings of individual images. Our analysis instead evaluates choices within a candidate set. \citet{seo2025} study cultural representation in image generation and editing; we study selection among existing outputs.

\citet{chang2026} train a cultural reward model on preferences derived from CulturalFrames and evaluate pairwise agreement. This is closely related evidence that cultural image preferences can be learned. We do not propose a competing reward model; we audit frozen, prompted selectors and the requests retained by their agreement gates.

\paragraph{VLM judges.}
\citet{zou2026} show that multimodal judges can favor informative answers without adequately using visual evidence. \citet{park2026} examine conflicts between textual narratives and perception, and train judges to ground their decisions more reliably. Both concern tasks distinct from our candidate-image selection setting. They motivate checking selected outcomes against independent ratings, but do not establish the mechanism behind the errors measured here.

K-Sort Eval uses multi-image VLM rankings, posterior correction, and adaptive comparisons to evaluate visual generators \citep{ksort2026}. Our study asks about the selected image within each prompt's candidate pool, rather than the accuracy or efficiency of a generator leaderboard. We do not claim that comparative image judging or best-of-pool selection is new.

\paragraph{Position bias in model judges.}
Language-model judges are sensitive to the order in which options are presented: pairwise judges favor a position \citep{zheng2023judging,wang2024fair,shi2025judging}, and multiple-choice accuracy shifts with option order \citep{pezeshkpour2024order}; judge verdicts also change when the judging instruction is reworded \citep{bellibatlu2026judgesense}. Multimodal judges also show position and consistency effects on vision-language benchmarks \citep{chen2024mllmjudge}, over-score earlier sentences and prefer their own captions when assessing image-text alignment \citep{saito2025haldec}, and can be swayed by visual manipulations unrelated to quality \citep{hwang2025fooling}. Proposed mitigations include swapping or averaging over orders \citep{zheng2023judging,wang2024fair} and de-biased judging protocols \citep{asaria2026judging}. VLM evaluators are also used to audit generated images at scale, with human validation on a subset \citep{seo2026portrait}. We measure the same phenomenon when the judge's output is an action, and ask how it interacts with agreement-based abstention.

\paragraph{Judgments used as decision rules.}
\citet{yao2026} investigate multimodal reinforcement learning with imperfect rewards, where gains on a proxy can diverge from task performance. WORLDVIEW examines cultural stereotyping introduced during prompt revision \citep{urman2026}. We isolate a later decision, selecting an existing candidate. The experiment does not optimize generators against a reward model or simulate a sequential feedback process.

\paragraph{Selective prediction.}
Risk--coverage analysis evaluates prediction with a reject option \citep{elyaniv2010,geifman2017}. We apply it to image selection, using agreement across orders or judges as gates. These gates yield measured operating points, not calibrated confidence scores or risk guarantees. Work on vision-language reasoning shows that selective models can abstain unnecessarily \citep{srinivasan2024recoverr}, and decision-theoretic metrics penalize confident errors more than abstentions \citep{wu2026bas}; our gate audit complements these by asking which decisions an agreement rule keeps and which it rejects.

\section{Problem and Measures}
\label{sec:measures}
\paragraph{Selection task.} Let $p$ denote a prompt and $C_p=\{x_{p1},\ldots,x_{pm_p}\}$ its pool of $m_p\in\{3,4\}$ candidate images, already generated and held fixed. A \emph{selection policy} $\pi$ receives the prompt and the images, shown in a given order with labels, and either returns the index $\pi(p)\in\{1,\ldots,m_p\}$ of one candidate or abstains. We always record the chosen image's identity, not its label, so the same image can carry the label ``A'' in one order and ``C'' in another.

\paragraph{Human reference utility.} Each candidate $x_{pi}$ has released prompt-alignment ratings from several annotators, each 0, 0.5, or 1. Its utility $h_{pi}\in[0,1]$ is the mean of these ratings. Judges never see these ratings; they are joined to the chosen image only after selection.

\paragraph{Baselines.} Choosing uniformly at random has expected utility $\bar h_p=\frac{1}{m_p}\sum_{i=1}^{m_p} h_{pi}$, which we compute exactly rather than by simulation. The \emph{annotation oracle} picks the highest-rated candidate, with utility $h^\star_p=\max_i h_{pi}$; it is a reference point, not a deployable selector.

\paragraph{Outcome measures.} For a policy that acts on prompt $p$ and returns image $x_{p,\pi(p)}$, with utility $h_{p,\pi(p)}$, we report three quantities:
\begin{align}
\text{regret:}\quad & R_p(\pi)=h^\star_p-h_{p,\pi(p)}, \label{eq:regret}\\
\text{gain over random:}\quad & G_p(\pi)=h_{p,\pi(p)}-\bar h_p, \label{eq:gain}\\
\text{below-mean event:}\quad & B_p(\pi)=\mathbf{1}\{h_{p,\pi(p)}<\bar h_p\}. \label{eq:bmr}
\end{align}
Regret is the primary outcome: zero means the judge returned a best-rated image. Gain is positive when the judge beats random choice on the same pool, and the two are linked by $G_p=(h^\star_p-\bar h_p)-R_p$. Averaging $B_p$ over prompts gives the \emph{below-mean rate} (BMR), a score shortfall relative to random, not a measure of user harm. Random choice has its own below-mean probability, the share of candidates rated below $\bar h_p$, which we use as its matched comparison. We also report how often the policy returns a best-rated image, counting ties. Every average is taken over the set $A_\pi$ of prompts on which the policy acts, and each policy is compared with random choice on exactly those prompts.

\paragraph{Cultural-error outcomes.} CulturalFrames also releases, per image, the share of annotators who flag a stereotype, a missing explicit expectation, or a missing implicit expectation. For each of these error rates we report the chosen image's value minus the random-choice average over the same pool, so that negative values mean fewer errors than chance. These outcomes are kept separate from alignment and are not combined into a composite score. A pool is \emph{sensitive} to an error type when that error rate differs across its candidates, so that selection can change it; we report results on each sensitive subset (Table~\ref{tab:decomposition}). As a secondary utility we also use overall satisfaction, rescaled from its 1--5 scale to $[0,1]$.

\paragraph{Presentation orders and agreement gates.} Each pool is shown in $K=3$ cyclic orders $k\in\{1,2,3\}$ (order~1 is canonical, orders~2 and~3 rotate it). An order places the candidates in labeled positions, which we call slots A, B, C, and D; a preference for particular slots regardless of content is \emph{position bias}. Let $\pi_k(p)$ be the index chosen under order $k$, and let $q_p$ be the share of the three orders that agree on the most frequent choice ($1/3$, $2/3$, or $1$). A \emph{gate} with threshold $\tau$ acts only when $q_p\ge\tau$ and then returns that majority choice: $\tau=1$ is \emph{unanimity}, $\tau=2/3$ requires two of three, and $\tau=1/3$ always acts. The \emph{cross-model gate} acts only when two judges make the same order-1 choice. For every gate we report coverage $|A_\pi|/N$ over the $N=300$ prompts, the outcomes on the prompts it keeps, and, because a gate can look good simply by keeping easy pools, the same outcomes for random choice and for the ungated judge on both the kept and the rejected prompts.

\paragraph{Country groups.} Each prompt belongs to one of ten country contexts; we report group means of regret and gain with their random baselines, and \emph{worst-group regret}, the largest group mean. Country labels identify dataset contexts, not homogeneous cultural preferences.

\section{Experimental Setup}
\subsection{Dataset and Frozen Splits}
We use the independently released CulturalFrames dataset \citep{culturalframes}. Image rows are joined to annotation records by exact image identifier, with prompt, country, and generator-suffix checks. We decode every image, check repeated identifiers and encoded-image content, and require at least three valid, annotated candidates. For repeated prompt text within a country we retain the lexicographically first base identifier before splitting, preventing duplicated requests from crossing splits. Missing annotations are never imputed. The provenance manifest pins the public dataset and model revisions.

The release contains 3,637 image rows and 3,577 image annotation records, comprising 10,412 individual ratings. 60 image rows have no annotation record (1.6\%). Filtering leaves 963 eligible sets. Removing 2 repeated requests within countries leaves 961 unique prompts: 313 three-candidate and 648 four-candidate sets. The main split has 100 three-candidate and 200 four-candidate sets; 11 sets have identical candidate alignment scores.

We create a development split of three prompts per country and a disjoint main split of 30 per country, using seed 20261002. Split membership depends only on a deterministic hash of prompt identifiers, without judge outputs or annotation values. The equal country allocation defines a balanced-country estimand; it does not reproduce a real deployment distribution. Remaining eligible sets form a reserve, and the main split is not expanded in response to its results.

\subsection{Open-Weight Judges}
The primary judge is Qwen3-VL-4B-Instruct (hereafter Qwen) \citep{qwenmodel}; SmolVLM2-2.2B-Instruct (Smol) is the secondary judge \citep{smolvlm,smolmodel}. Model cards, revisions, dependency versions, and inference settings are stored with the artifact. We run locally with greedy decoding, a 64-token output limit, and no requested explanation. Qwen images fit an aspect-preserving 448-pixel bounding box. Smol uses its native 384-pixel global view with image tiling disabled to bound memory consumption. The two judges therefore receive different visual token budgets, so differences between them cannot be attributed to model capability alone.

Every request interleaves a candidate label with its image. The instruction asks for explicit prompt satisfaction, implicit contextual consistency, avoidance of unsupported stereotypes, and visual integrity. It warns against rewarding cultural symbols merely for their number or recognizability. Only the original prompt supplies cultural context. Source model names, category labels, country metadata, and human annotations do not enter the request.

For Qwen the output schema is a JSON object with \texttt{choice} and \texttt{ranking}. The ranking must contain every supplied label exactly once, and its first label must equal the choice. Smol receives the same evaluation criteria but a choice-only JSON contract, selected during development because it could not reliably satisfy the full-ranking format. Its adapter can normalize a single explicit label in a one-field structured object; raw JSON and exact-schema validity are tracked separately from recovered decision validity. Rankings are not used by any outcome metric. These interface differences limit comparisons of formatting reliability. We record raw text, parsed output, label-to-image mapping, input tensor shapes, model revision, elapsed time, and request hashes. Unit tests verify regret extrema, the analytic random expectation, tie behavior, annotation aggregation, and identity preservation under permutation.

\subsection{Candidate Order and Invalid Outputs}
Each prompt receives three distinct cyclic presentations: its canonical identifier order and two deterministic rotations. Canonical order sorts candidate identifiers, which usually places the same generator first; the rotations break this link, so generator effects are assessed over all three orders. The instruction, images, and decoding settings remain fixed. Thus disagreement measures sensitivity to these presentation changes rather than sampled decoding variation or prompt paraphrases. Three rotations do not exhaust all orders, especially for four candidates.

The original-order decision is the primary direct policy. No main-run prompt edits are permitted. An unresolved parse failure produces an abstention and remains in the output log. Qwen uses strict schema validation; Smol uses its predeclared single-label normalization, with every recovery flagged. A runtime exception permits one retry with identical inputs; invalid text is not repeatedly sampled until valid. Utility and regret are conditional on valid decisions, with coverage relative to the full frozen split reported alongside them. We never assign failed requests a favorable candidate or silently remove their denominator.

\subsection{Selection with Abstention}
The gate thresholds $\tau\in\{1/3,2/3,1\}$ and the cross-model rule were declared before main inference. A gate acts only when all three order calls parse; ties in the majority are broken by canonical identifier order, so the action never depends on human ratings. Abstention leaves a request unresolved: we do not assume a human fallback or a zero-risk outcome, and we report coverage next to every selective estimate. Because Smol is a weaker judge (Section~\ref{sec:results}), cross-model results are conditional on that pairing.

\subsection{Statistical Analysis}
The unit of analysis is a prompt with its candidate pool; the three order calls for a prompt are never resampled independently. We report 95\% intervals from 10,000 prompt-level bootstrap resamples (seed 20261002) for means and paired differences, and Wilson intervals for a single policy's BMR. Intervals condition on the released annotation means. Secondary cultural-error intervals are pointwise, and we additionally report Bonferroni-adjusted intervals across the three error types. Analyses of presentation position, of the prompts that gates reject, and all ablations and the scale extension in Sections~\ref{sec:ablations} and~\ref{sec:scale} were added after main inference and are labeled exploratory. Appendix~\ref{app:stats} gives further estimation details.

\section{Results}
\label{sec:results}

\begin{table*}[t]
\centering
\resizebox{\textwidth}{!}{\begin{tabular}{lrrcllrrc}
\toprule
Policy & $n$ & Cov.\,(\%) & Align. & Regret [95\% CI] & Gain over random [95\% CI] & BMR\,(\%) & Best\,(\%) & Worst\\
\midrule
Random & 300 & 100.0 & 0.730 & 0.199 [0.186, 0.213] & 0.000 [0.000, 0.000] & 43.1 & 41.9 & 0.266\\
Qwen & 300 & 100.0 & 0.770 & 0.160 [0.136, 0.184] & 0.039 [0.016, 0.062] & 36.3 & 51.0 & 0.275\\
Qwen, majority of 3 orders & 300 & 100.0 & 0.795 & 0.135 [0.114, 0.156] & 0.065 [0.043, 0.086] & 31.7 & 54.3 & 0.186\\
Qwen, unanimity gate (3/3 orders) & 121 & 40.3 & 0.854 & 0.074 [0.050, 0.100] & 0.141 [0.110, 0.173] & 16.5 & 71.1 & 0.115\\
Smol & 297 & 99.0 & 0.659 & 0.270 [0.243, 0.298] & $-$0.073 [$-$0.093, $-$0.052] & 60.3 & 29.0 & 0.400\\
Cross-model gate & 59 & 19.7 & 0.690 & 0.249 [0.187, 0.316] & $-$0.060 [$-$0.111, $-$0.011] & 54.2 & 33.9 & 0.490\\
CLIP baseline & 300 & 100.0 & 0.799 & 0.130 [0.109, 0.153] & 0.069 [0.047, 0.090] & 28.0 & 55.7 & --\\
Oracle & 300 & 100.0 & 0.929 & 0.000 [0.000, 0.000] & 0.199 [0.186, 0.213] & 0.0 & 100.0 & 0.000\\
\bottomrule
\end{tabular}
}
\caption{Selection outcomes on the 300 main prompts (95\% prompt-level bootstrap intervals). Cov.: share of prompts on which the policy acts. Align.: mean prompt-alignment rating of returned images. Gain over random is paired with exact random choice on the same prompts. BMR: share of returned images rated below the pool mean. Best: share of best-rated returns, ties included. Worst: highest country-group mean regret. Gated policies are evaluated on the prompts they keep. CLIP is a content-only similarity scorer (Section~\ref{sec:clip}).}
\label{tab:main}
\end{table*}

\begin{figure*}[t]
\centering
\includegraphics[width=\textwidth]{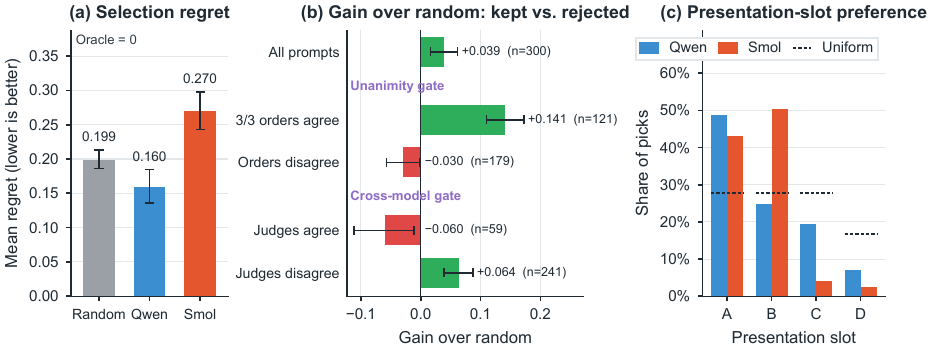}
\caption{Main results (95\% prompt-level bootstrap intervals). (a) Selection regret; lower is better. (b) Gain over random choice on the prompts each gate keeps and on those it rejects. Order unanimity keeps prompts where Qwen beats random and rejects prompts where it falls below random; agreement between the two judges does the reverse. (c) Share of choices by presentation slot; dashed marks show the rate expected under uniform choice given the pool sizes.}
\label{fig:results}
\end{figure*}

\subsection{Direct Selection}
Table~\ref{tab:main} and Figure~\ref{fig:results}a summarize the direct policies. Qwen returns a valid original-order choice for all 300 prompts. Its mean alignment is 0.770, against 0.730 for random choice and 0.929 for the oracle, a gain of 0.039 (95\% CI [0.016, 0.062]). Mean regret is 0.160 [0.136, 0.184]; Qwen returns a best-rated image for 51.0\% of prompts and a below-mean image for 36.3\%, against 43.1\% for random choice (paired difference $-$0.067, 95\% CI [$-$0.118, $-$0.016]).

Smol does worse than random: gain $-$0.073 [$-$0.093, $-$0.052], regret 0.270, and BMR 60.3\%. It rarely follows the output format (Appendix Table~\ref{tab:validity}), and its interface and image resolution differ from Qwen's, so we treat it as a weak secondary judge rather than as a capability comparison. Appendix Table~\ref{tab:sensitivity} shows that the conclusions hold with a below-mean margin of 0.05, with the median in place of the mean, and with overall satisfaction as the utility.

\subsection{Presentation Position}
Both main judges favor early slots (Figure~\ref{fig:results}c). Qwen picks slot A in 48.8\% of its 900 calls, where uniform choice given the pool sizes would give 27.8\% ($\chi^2_{3}=219$, $p<10^{-10}$); Smol places 93.3\% of its choices in slots A or B. Consistent with this, Qwen's choice changes across the three orders for 59.7\% of prompts. Figure~\ref{fig:hero} shows the extreme case: the judge returns whatever image is shown first.

The judge still responds to content. Canonical order places one generator first in most pools, so we pool all three rotations, which spreads each generator across slots: Qwen then chooses images from the generator with the highest mean human rating in 57.9\% of calls, against 27.2\% under uniform choice (Appendix Table~\ref{tab:generators}). Its choices mix a content signal with a strong slot preference.

\subsection{What Agreement Gates Keep and Reject}
A judge that follows position alone picks a different image in each rotation, so it can never be unanimous. Unanimity keeps 121 prompts (40.3\%). On them Qwen's gain over random is 0.141 [0.110, 0.173] and regret 0.074; on the 179 rejected prompts its gain is $-$0.030 [$-$0.057, $-$0.002], below random (Figure~\ref{fig:results}b). The kept prompts are not easier: their candidates differ more in rating than the rejected ones (mean best-minus-worst range 0.454 versus 0.404), leaving more room for error. The gate only filters prompts; on kept prompts the choice is unchanged.

The two-of-three gate is an intermediate operating point: it keeps 80.7\% of prompts with gain 0.094 [0.071, 0.118] (Appendix Table~\ref{tab:policies}).

Voting over the three orders instead of abstaining, a simple form of order averaging, raises the gain on all prompts from 0.039 to 0.065 [0.043, 0.086].

Agreement between judges behaves differently. Qwen and Smol agree on 59 prompts (19.7\%), where Qwen's gain is $-$0.060 [$-$0.111, $-$0.011]; on the 241 prompts where they disagree it is 0.064 [0.039, 0.087]. Two judges that both prefer early slots can agree for reasons unrelated to content, so this gate keeps the wrong prompts. The result is specific to pairing with a weak judge. Appendix Table~\ref{tab:policies} lists every gate with its complement.

\subsection{Comparison with a CLIP Baseline}
\label{sec:clip}
Would a simpler scorer that cannot be influenced by order do as well? We score each candidate by CLIP prompt--image similarity \citep{radford2021clip,hessel2021clipscore} and return the highest-scoring image. This CLIP baseline has regret 0.130 (95\% CI [0.109, 0.153]) and gain over random 0.069 [0.047, 0.090], compared with Qwen's 0.160 and 0.039; the paired difference in gain favors CLIP by 0.030 [0.005, 0.055]. CLIP also selects a below-mean image less often (28.0\% versus 36.3\%). Yet it raises the stereotype rating relative to random by 0.026 [0.006, 0.046], as much as Qwen does, while sharply reducing missing explicit expectations ($-$0.113). An alignment-oriented scorer therefore does not avoid stereotyped images; this concern is not specific to VLM judges.

Qwen agrees with CLIP on 49.0\% of prompts (chance 27.8\%). Where they agree, Qwen's gain is 0.099 [0.067, 0.130]; where they disagree it is $-$0.018 [$-$0.047, 0.011]. Agreement with an independent, position-invariant scorer thus behaves like order unanimity rather than like agreement with the weaker VLM judge. The CLIP comparison is exploratory (Appendix Table~\ref{tab:baselines}).

Results do not hinge on noisy annotations. Splitting prompts at the median annotator disagreement, Qwen's gain is 0.047 [0.014, 0.079] on high-consensus and 0.033 [0.002, 0.062] on low-consensus prompts (Appendix Table~\ref{tab:robustness}).

The CLIP baseline embeds the raw prompt (no template; all prompts fit the 77-token limit) and each image with the default CLIP preprocessing (resize and center crop to 224 pixels), and returns the image with the highest cosine similarity; ties follow canonical order. Our main baseline uses ViT-L/14. A smaller ViT-B/32 backbone gives nearly the same gain (0.063 [0.042, 0.085]; difference $-$0.006 [$-$0.028, 0.018]), but no stereotype increase (0.003 [$-$0.016, 0.023]), so the stereotype effect of a similarity scorer depends on the backbone (Appendix Table~\ref{tab:extras-clip}).

\subsection{Is the Position Bias an Artifact?}
\label{sec:ablations}
\begin{figure}[t]
\centering
\includegraphics[width=\columnwidth]{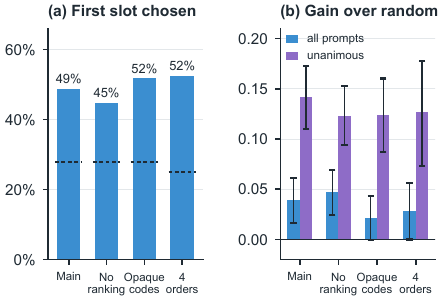}
\caption{Post-hoc ablations with Qwen. (a) Share of calls choosing the first-presented image; dashes mark the uniform rate. (b) Gain over random on all prompts and on prompts where all orders agree (95\% intervals). No ranking: choice-only output format. Opaque codes: two-character labels instead of letters. 4 orders: a fourth rotation for four-image pools.}
\label{fig:ablations}
\end{figure}

The position bias could be an artifact of our interface rather than of the judge. We ran three post-hoc ablations with Qwen on all 300 prompts, leaving the frozen main results unchanged (Figure~\ref{fig:ablations}, Appendix Table~\ref{tab:ablations}).

\textit{Output format.} The main contract asks for a full ranking whose first entry is the choice, which could prime the first label. With a choice-only contract, Qwen still picks the first slot in 44.8\% of calls (uniform 27.8\%) and changes its choice across orders on 54.0\% of prompts. Its gain over random is 0.047 [0.024, 0.069], and unanimity keeps 46.0\% of prompts with gain 0.123 [0.094, 0.153].

\textit{Label identity.} Replacing A/B/C/D with opaque two-character codes unrelated to order, Qwen picks the first-presented image in 51.8\% of calls (uniform 27.8\%) and changes its choice on 62.7\% of prompts, with gain 0.021 [$-$0.001, 0.044].

\textit{Full rotation.} Adding a fourth rotation for the 200 four-image pools places every image in every slot once. Slot shares are then A 52.5\%, B 20.4\%, C 14.4\%, D 12.8\% (uniform 25.0\%). Unanimity over four orders keeps 27.5\% of these pools with regret 0.095 and gain 0.127 [0.073, 0.178], and voting over the four orders gives gain 0.061 [0.034, 0.087].

In every variant the bias toward the first slot remains and unanimity still keeps prompts on which the judge beats random. The preference is a property of the judge, not of the letter labels or the ranking format.

\subsection{Does Scale Remove the Bias?}
\label{sec:scale}
\begin{figure}[t]
\centering
\includegraphics[width=\columnwidth]{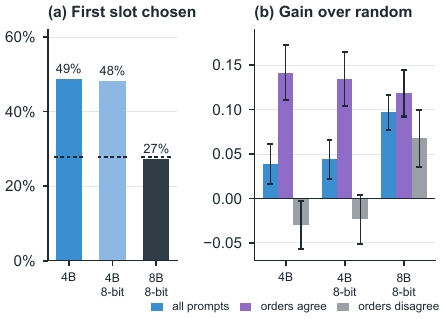}
\caption{Scale extension (8B). (a) Share of calls choosing the first slot; dashes mark the uniform rate. (b) Gain over random on all prompts, on prompts where all three orders agree, and on prompts where they do not (95\% intervals). The 8-bit 4B run is a control for quantization.}
\label{fig:scale}
\end{figure}

Is position bias a property of small judges? We ran the larger Qwen3-VL-8B-Instruct on the same 300 prompts and three orders. To fit it in memory we used 8-bit weights, so we also reran the 4B judge with the same quantization as a control. The control agrees with the main 4B run on 94.9\% of calls and reproduces its first-slot share (48.2\%) and gain (0.044), so neither quantization nor the MLX runtime explains the differences below (Figure~\ref{fig:scale}, Appendix Table~\ref{tab:scale}).

At 8B the position bias largely disappears: the first slot is chosen in 27.3\% of calls, close to the uniform 27.8\% ($p=0.06$), although the choice still changes across orders on 43.0\% of prompts. The 8B judge is also much better: its gain over random is 0.097 [0.077, 0.117], above the CLIP baseline (0.069; paired difference 0.028 [0.003, 0.054]), and its stereotype difference is no longer detectable (0.016 [$-$0.004, 0.036]).

The value of the unanimity gate changes with it. For the 4B judge, the prompts unanimity rejects fall below random; for the 8B judge they stay above random (0.068 [0.036, 0.099]), so the gate mostly discards good decisions. Agreement between the 8B and 4B judges shows the same pattern: on the 126 prompts where they disagree, the 8B choice has gain 0.098 and the 4B choice $-$0.039. A gate is useful only while it filters the failure mode of the judge in front of it, and must be re-audited when the judge changes.

\subsection{Cultural-Error Outcomes}
\begin{figure}[t]
\centering
\includegraphics[width=\columnwidth]{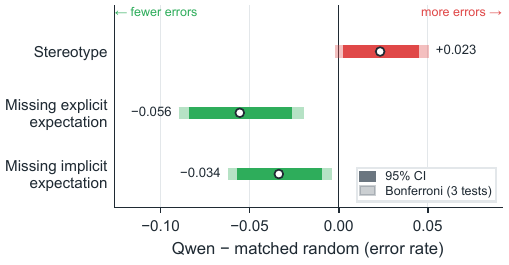}
\caption{Cultural-error ratings of Qwen's choices minus random choice on the same pools; negative values mean fewer errors. Dark bars: pointwise 95\% intervals. Light bars: Bonferroni-adjusted intervals across the three error types. Qwen reduces missing expectations and slightly raises stereotype ratings; the stereotype difference does not survive the adjustment.}
\label{fig:cultural}
\end{figure}

Figure~\ref{fig:cultural} compares the cultural-error ratings of Qwen's choices with random choice on the same pools. Qwen lowers missing explicit expectations by 0.056 [$-$0.084, $-$0.026] and missing implicit expectations by 0.034 [$-$0.057, $-$0.009], and both reductions survive a Bonferroni adjustment across the three error types. It raises the stereotype rating from 0.125 to 0.148 (difference 0.023, pointwise 95\% CI [0.002, 0.045]); this interval includes zero after adjustment ([$-$0.002, 0.051]), so we treat it as exploratory. On the pools where stereotype ratings differ across candidates, the difference is 0.036 [0.004, 0.068] (Appendix Table~\ref{tab:decomposition}). An alignment gain therefore need not carry over to every cultural outcome.

Where does the stereotype increase come from? Most of it traces to one generator: Qwen returns SD-3.5-Large images in 41.3\% of prompts against 27.6\% under uniform choice, and these images carry the highest stereotype ratings (Appendix Table~\ref{tab:extras-stereo}). The increase is also larger when several candidates tie for the best alignment (0.038 [0.004, 0.073]) than when one candidate is clearly best (0.014 [$-$0.012, 0.040]). Aggregating across orders leaves no detectable increase: the stereotype difference is 0.007 [$-$0.013, 0.027] for voting over the three orders and 0.010 [$-$0.019, 0.039] on unanimous prompts, whereas on the prompts unanimity rejects it is 0.032 [0.003, 0.063]. Both aggregated policies still reduce missing explicit expectations ($-$0.092 and $-$0.174).

\subsection{Country Groups}
Qwen's point estimate of gain is positive in 7 of 10 country groups (Appendix Figure~\ref{fig:country}), ranging from 0.107 in Canada to $-$0.018 in India. Regret is highest in Japan (0.275, pointwise 95\% CI [0.171, 0.387]), where random choice also has the highest regret (0.266). With 30 prompts per group, these intervals are wide and overlap; country labels describe dataset contexts, not cultural preferences, and the analysis is not powered to rank groups.

\section{Discussion}
\paragraph{Evaluate the action, not the score.}
A judge that ranks images plausibly can still return a poor image when it must act. Comparing the returned image with the best available one and with random choice on the same pool makes the value of the decision visible, and it exposes failures that score correlation hides: here, a strong dependence on where an image appears.

\paragraph{Agreement gates need their own audit.}
Agreement is useful only when it breaks the judge's failure mode. Agreement can retain stable mistakes, and any reduction in regret must be compared with random choice on the kept prompts, and with the judge on the rejected ones. Position bias explains why the two gates we studied diverge. Unanimity across rotations demands the same image from different slots, so it filters out position-driven choices; two judges with similar slot preferences can agree for reasons unrelated to content. Order averaging, by voting across rotations, recovers part of the lost value without abstaining. Abstention still leaves a practical question: who handles the rejected request, at what cost, and with what outcome.

\paragraph{Is the unanimity result explained by easy prompts?}
No. Candidates on the prompts that unanimity keeps differ more in rating than those on the rejected prompts, so the kept pools leave more room for error, not less (Section~\ref{sec:results}); the judge's gain on them is large, and its gain on the rejected prompts is below zero.

\paragraph{What changes with scale?}
The 8B extension shows that the specific failure is not universal: position bias nearly vanishes, and with it the gate's ability to screen out below-random decisions, although order sensitivity itself remains (Section~\ref{sec:scale}). The audit itself transfers unchanged and is what reveals the difference. This is the practical lesson: a filter validated for one judge is not validated for the next.

\paragraph{What order checks cost.}
Order checks are cheap relative to their value. On our hardware a single Qwen decision takes a median of 3.5~s and three orders 10.6~s, roughly three times the cost, which buys either the voting gain or the unanimity filter. Adding a fourth order lowers coverage without lowering regret further (four-order unanimity keeps 27.5\% of four-image pools at regret 0.095) (Appendix Figure~\ref{fig:risk-coverage}).
 Order sensitivity is only weakly related to how close the candidates are: Qwen's choice changes across orders on 62.8\% of prompts in the lowest third of rating range and 53.6\% in the highest (Appendix Table~\ref{tab:extras-difficulty}), so the bias is not confined to near-ties.

\paragraph{Alignment and cultural errors are different targets.}
Both the VLM judge and the CLIP baseline reduce missing cultural expectations, but stereotype ratings move differently: the judge raises them slightly, and so does one CLIP backbone but not another. Aggregating the judge's decisions across orders leaves no detectable increase. A selector tuned or validated on prompt alignment should therefore be audited separately on stereotype outcomes, and composite scores would hide this trade-off.

\section{Limitations and Ethics}
CulturalFrames covers a limited set of countries, domains, prompts, and generators. Country labels compress substantial within-country diversity. Candidate images are existing outputs from a small generator pool, so the results depend on its visual styles and available alternatives. Hiding generator names does not remove recognizable stylistic signatures. Our local input resolution can obscure fine details and is part of the evaluated configuration.

Human annotation means are imperfect references. Different annotators can disagree, small rating panels create ties and noise, and selecting a maximum can exploit annotation noise. The oracle should be interpreted accordingly. Primary bootstrap intervals condition on these means; they are not estimates of uncertainty over all possible human evaluations. Prompts can share semantic templates even after exact-text deduplication, so unmodeled dependence may also narrow intervals. Prompt alignment reflects ordinary request satisfaction and cannot isolate culture-specific value by itself. Missing-expectation and stereotype endpoints provide additional descriptions without an arbitrary composite.

The setting simulates selecting a returned image; it measures neither observed user harm nor live sequential editing. We do not treat independent generations as temporal trajectories or claim a feedback-amplification result. Our main judges are small (4B and 2.2B parameters); a single 8B extension, run with 8-bit weights and a matched quantization control, shows that position bias can shrink sharply with scale, and we do not test larger or proprietary models. With 30 prompts per country, country-group intervals are wide, and the study is underpowered to detect moderate group differences. Position, complement-set, content-baseline, ablation, and scale analyses were added after main inference and are exploratory. The CLIP comparison uses two backbones of one similarity model family. Agreement statistics are not epistemic confidence, and selective risk is not overall system risk when abstained cases remain unresolved.

This is secondary analysis of an independently released public benchmark, with no new participant recruitment or collection of personal data. Derived artifacts retain aggregate scores and candidate identifiers, not annotator identifiers, demographics, or free-text comments. The public dataset card does not declare a dataset license; public availability does not establish redistribution permission. We provide newly written code, manifests, and derived outputs; source images and raw human records are not redistributed. Figures~\ref{fig:hero}, \ref{fig:teaser}, and~\ref{fig:gallery} reproduce dataset images with attribution for illustration. Applicable institutional requirements for secondary analysis remain an author responsibility.

\paragraph{AI assistance.}
AI coding and writing assistants (OpenAI Codex and Anthropic Claude) assisted with code development, manuscript drafting and revision, figure code, and draft review. All empirical results are computed from released annotations and recorded local judge outputs; writing assistance does not supply experimental observations. Authors remain responsible for verifying the analysis, references, and final manuscript.

\paragraph{Statement of Contributions.}
This paper contributes (i) a matched-baseline, order-rotated audit protocol for VLM judges that select from fixed candidate pools, released with a deterministic artifact and an offline browser for every recorded decision (Appendix~\ref{app:browser}); (ii) evidence of strong position bias in two small open judges, with post-hoc ablations of the output format and label identity and a scale extension showing that the bias shrinks in a larger judge; and (iii) an audit of agreement gates on both the prompts they keep and the prompts they reject, compared with a content-only baseline. The paper has no earlier published version; all experiments, analyses, and text are new.

\section{Conclusion}
A 4B VLM judge that chooses which image to return barely beats random choice (0.039), falls short of a CLIP similarity scorer, and picks the first-listed image in 48.8\% of calls. Agreement across orders separates its good decisions from its bad ones; agreement with a weaker judge that shares its position bias does not. A larger 8B judge from the same family shows almost no position bias, and for it the same order-agreement gate mostly discards good decisions; a gate validated for one judge is not validated for the next. Agreement is useful only when it targets the failure mode of the judge in front of it, so selection audits, including audits of the filters around a judge, must be repeated when the judge changes.

These findings concern a single choice from a fixed candidate pool. Evaluating a deployed selector also requires evidence about its users, later actions, and outcomes when it abstains.

\bibliography{refs}

\appendix
\section{Reproduction Protocol}
Seed: 20261002. Dataset revision: \texttt{7a44117634da}. Qwen revision: \texttt{ebb281ec70b0}. Smol revision: \texttt{482adb537c02}. Full revision identifiers and runtime versions are in the provenance manifest. The main split contains 300 prompts, 30 per country; the dev split contains 30 disjoint prompts. Each prompt has three cyclic candidate orders. Qwen uses a 448-pixel maximum image edge; Smol uses one 384-pixel native view. Both use greedy decoding and a 64-token limit. The image audit finds 0 undecodable images and 0 identifier/content mismatches. Four repeated encoded-image pairs occur across the duplicate requests removed before splitting.

In the artifact, the key \texttt{hsr} stores the below-mean rate (BMR) reported here. The pipeline is deterministic given the recorded outputs: \texttt{make analyze} recomputes every table, figure, and number in this paper from the frozen judge logs, and \texttt{make paper} regenerates the manuscript text fragments. Figures and prose never contain hand-entered estimates; each value is read from the analysis outputs at build time, and a checksum manifest covers the annotations, manifests, judge logs, and analysis code.

\section{Additional Result Figures and Tables}
\begin{figure*}[t]
\centering
\includegraphics[width=\textwidth]{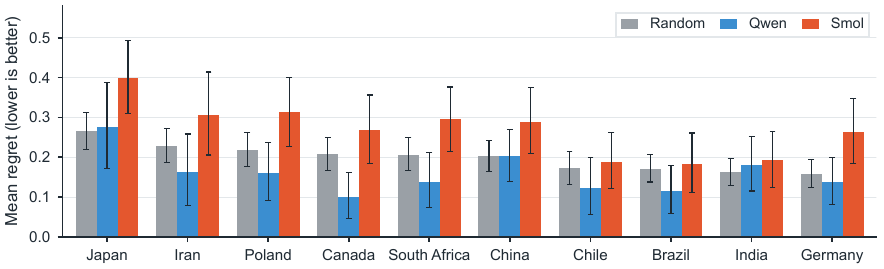}
\caption{Country-group regret with 95\% bootstrap intervals (30 prompts per group). Random regret shows each group's room for improvement.}
\label{fig:country}
\end{figure*}

\begin{figure}[t]
\centering
\includegraphics[width=\columnwidth]{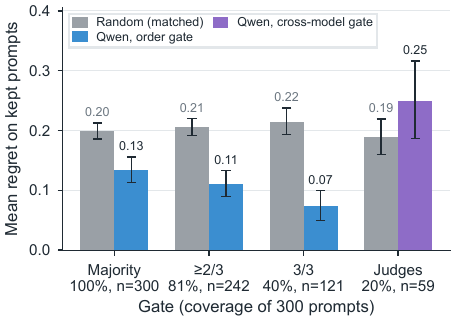}
\caption{Regret at each agreement gate next to random choice on the identical kept prompts (95\% bootstrap intervals). ``Majority'' votes over the three orders and always acts.}
\label{fig:coverage}
\end{figure}

Figures~\ref{fig:country} and~\ref{fig:coverage} expand the country and agreement-gate results, and Table~\ref{tab:decomposition} gives results on the error-sensitive subsets.
\label{app:figs}

\begin{table*}[t]
\centering
\footnotesize
\begin{tabular*}{\textwidth}{@{\extracolsep{\fill}}lrrrrl}
\toprule
Condition & $n$ & Regret & Selected & Random & Difference [95\% CI]\\
\midrule
Explicit & 276 & 0.169 & 0.280 & 0.341 & $-$0.060 [$-$0.092, $-$0.029]\\
Implicit & 248 & 0.168 & 0.207 & 0.248 & $-$0.041 [$-$0.068, $-$0.012]\\
Stereotype & 194 & 0.169 & 0.227 & 0.191 & 0.036 [0.004, 0.068]\\
\bottomrule
\end{tabular*}

\caption{Qwen's choices on pools where a given error rating differs across candidates. Selected and Random are mean error rates; the difference is paired (95\% CI). Conditions overlap and do not isolate causal effects.}
\label{tab:decomposition}
\end{table*}

\section{Dataset Composition}
Table~\ref{tab:dataset} describes the frozen main split by country. \emph{Ties} are pools with identical alignment scores (zero regret under every policy); \emph{Mean range} is the average spread between the best and worst candidate, an upper bound on the regret a judge can incur. Table~\ref{tab:generators} reports each generator's representation and mean rating.

\begin{table*}[t]
\centering

\footnotesize
\begin{tabular*}{\textwidth}{@{\extracolsep{\fill}}lrrrrrrr}
\toprule
Country & Prompts & 3-img & 4-img & Ties & Mean align. & Mean range & Stereo.\\
\midrule
Brazil & 30 & 11 & 19 & 1 & 0.790 & 0.410 & 0.109\\
Canada & 30 & 7 & 23 & 1 & 0.741 & 0.454 & 0.044\\
Chile & 30 & 8 & 22 & 3 & 0.762 & 0.378 & 0.069\\
China & 30 & 14 & 16 & 1 & 0.690 & 0.389 & 0.154\\
Germany & 30 & 13 & 17 & 2 & 0.790 & 0.339 & 0.100\\
India & 30 & 7 & 23 & 0 & 0.763 & 0.369 & 0.091\\
Iran & 30 & 8 & 22 & 1 & 0.700 & 0.514 & 0.264\\
Japan & 30 & 10 & 20 & 1 & 0.596 & 0.503 & 0.218\\
Poland & 30 & 10 & 20 & 0 & 0.714 & 0.467 & 0.106\\
South Africa & 30 & 12 & 18 & 1 & 0.757 & 0.418 & 0.078\\
\bottomrule
\end{tabular*}

\caption{Main split by country (30 prompts each). Mean align. and Stereo. average candidate-level human ratings.}
\label{tab:dataset}
\end{table*}

\begin{table*}[t]
\centering

\footnotesize
\begin{tabular*}{\textwidth}{@{\extracolsep{\fill}}lrrrrr}
\toprule
Generator & Images & Mean align. & Uniform (\%) & Qwen (\%) & Smol (\%)\\
\midrule
Flux.1-Dev & 296 & 0.639 & 27.3 & 8.4 & 35.2\\
GPT-Image & 295 & 0.854 & 27.2 & 57.9 & 30.2\\
Imagen3 & 211 & 0.785 & 17.9 & 16.1 & 9.3\\
SD-3.5-Large & 298 & 0.660 & 27.6 & 17.6 & 25.3\\
\bottomrule
\end{tabular*}

\caption{Generators in the main split. Uniform is the share expected if a judge chose uniformly at random given pool sizes; Qwen and Smol are choice shares pooled over all three orders, which decouples generator from presentation slot.}
\label{tab:generators}
\end{table*}

\section{Additional Qualitative Examples}
\label{app:gallery}
\begin{figure*}[p]
\centering
\includegraphics[width=0.86\textwidth]{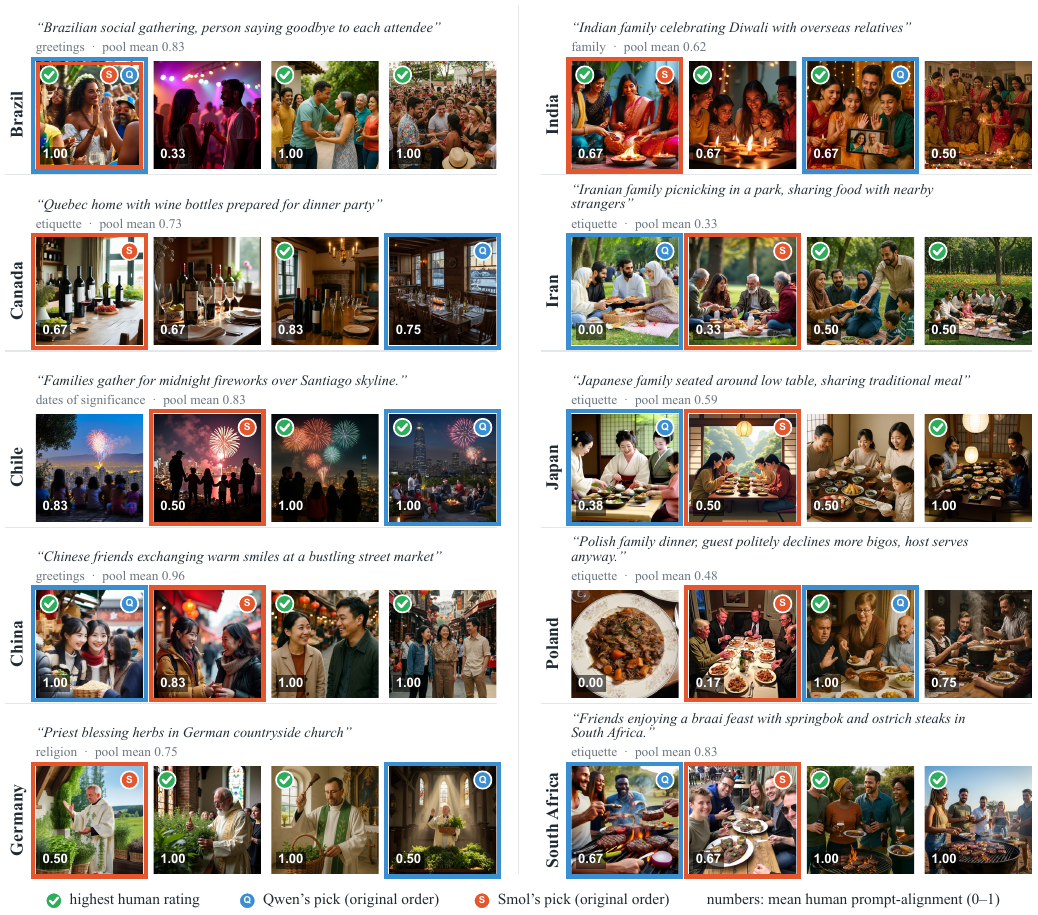}
\caption{Unfiltered examples: the first four-candidate prompt per country in the frozen manifest, with candidates in original presentation order. Numbers are mean human prompt-alignment ratings. Blue (Q) and orange (S) frames mark the Qwen and Smol original-order choices; green checks mark the highest-rated candidates. Images from CulturalFrames \citep{culturalframes}.}
\label{fig:gallery}
\vspace{1.4em}
\includegraphics[width=0.9\textwidth]{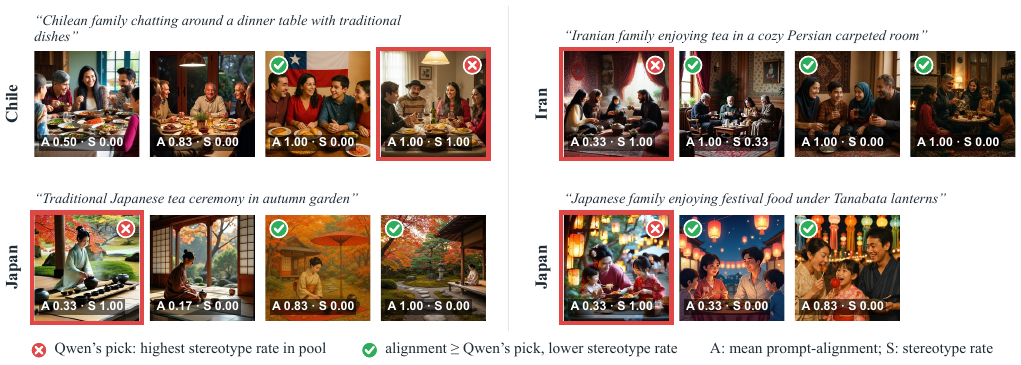}
\caption{Stereotype trade-off. Prompts where Qwen's choice (red) has the highest stereotype rate in its pool and another candidate with equal or higher alignment has a lower rate (checks). Labels give mean alignment (A) and the share of annotators flagging a stereotype (S). Rows with three images are prompts whose pool contains only three candidates.}
\label{fig:gallery-stereo}
\end{figure*}

Figure~\ref{fig:teaser} uses a fixed rule applied without viewing images: among three-candidate pools where Qwen's original-order choice is below the pool mean and some candidate exceeds the mean by at least 0.25, take the largest-regret prompt per country, then the four countries with the largest regret (ties broken by name). Figure~\ref{fig:gallery} shows one pool per country, chosen as the first four-candidate prompt in the frozen manifest. Unlike Figure~\ref{fig:teaser}, this sample is not filtered on outcomes. Both judges' original-order choices are marked.

\begin{figure*}[p]
\centering
\includegraphics[width=0.9\textwidth]{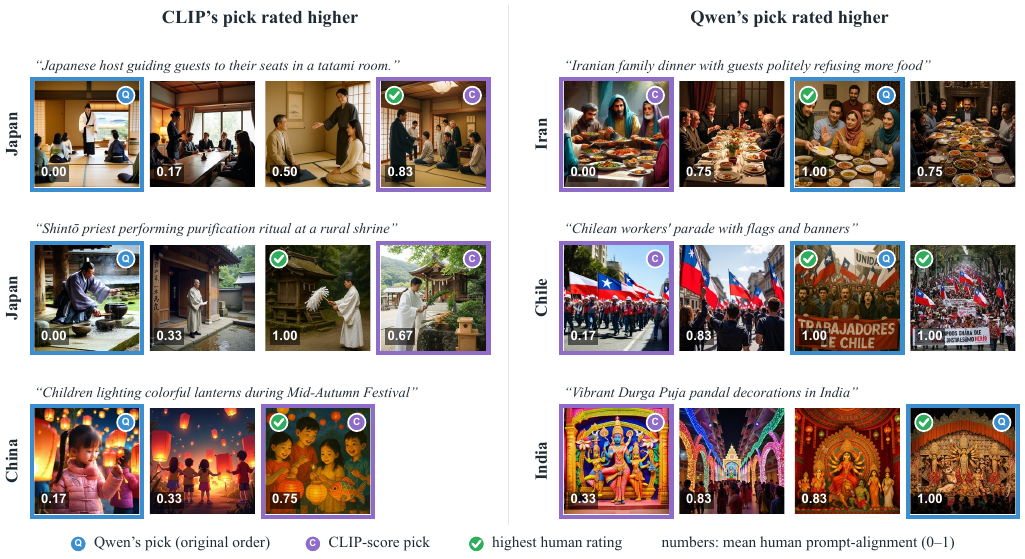}
\caption{Content scorer versus judge. Prompts where CLIP's choice (C, purple) and Qwen's original-order choice (Q, blue) differ most in rating: left, CLIP's choice is rated higher; right, Qwen's is. Checks mark the best-rated candidates. Rows with three images are prompts whose pool contains only three candidates.}
\label{fig:gallery-clip}
\vspace{1.4em}
\includegraphics[width=0.72\textwidth]{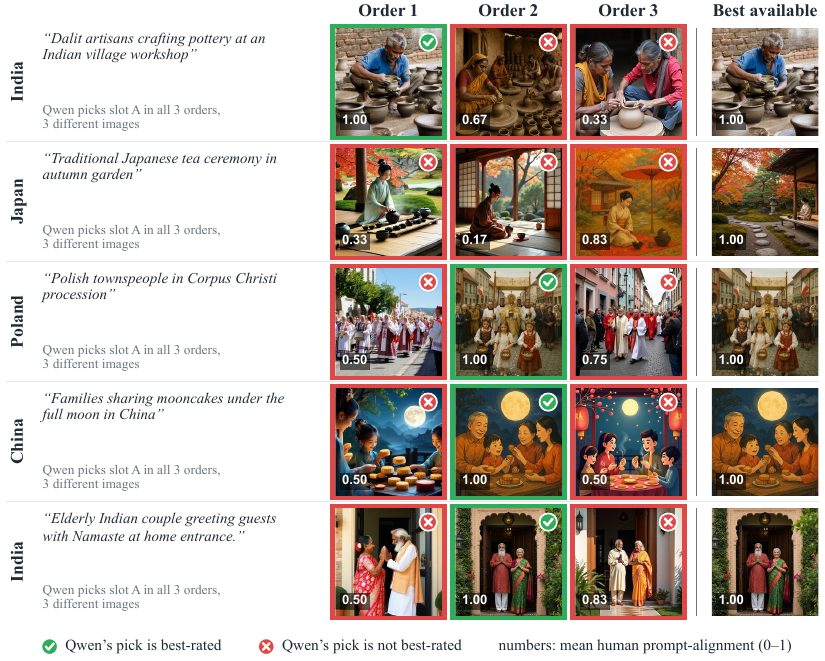}
\caption{Order flips. Prompts where Qwen picks slot A in all three orders and returns at least two different images; the five with the widest rating range among its picks. The last column is the best-rated candidate. Images from CulturalFrames \citep{culturalframes}.}
\label{fig:gallery-orders}
\end{figure*}

Three further galleries use fixed selection rules documented in the artifact. Figure~\ref{fig:gallery-orders} shows prompts on which Qwen picks slot A in every order, so the returned image changes with order alone. Figure~\ref{fig:gallery-clip} contrasts Qwen with the CLIP scorer where they disagree most, in both directions. Figure~\ref{fig:gallery-stereo} shows prompts where Qwen returns the most stereotyped image in the pool although an equally or better aligned image with a lower stereotype rate was available.

\begin{table*}[p]
\centering

\footnotesize
\begin{tabular*}{\textwidth}{@{\extracolsep{\fill}}lrrllrrr}
\toprule
Policy & $n$ & Cov. & Regret [95\% CI] & Gain [95\% CI] & BMR & Rand. BMR & Best\\
\midrule
Random & 300 & 100.0 & 0.199 [0.186, 0.213] & 0.000 [0.000, 0.000] & 43.1 & 43.1 & 41.9\\
Qwen & 300 & 100.0 & 0.160 [0.136, 0.184] & 0.039 [0.016, 0.062] & 36.3 & 43.1 & 51.0\\
Qwen majority & 300 & 100.0 & 0.135 [0.114, 0.156] & 0.065 [0.043, 0.086] & 31.7 & 43.1 & 54.3\\
Qwen agree 2/3 & 242 & 80.7 & 0.111 [0.090, 0.133] & 0.094 [0.071, 0.118] & 24.8 & 43.0 & 61.2\\
Qwen unanimous & 121 & 40.3 & 0.074 [0.050, 0.100] & 0.141 [0.110, 0.173] & 16.5 & 43.5 & 71.1\\
Qwen, not unanimous & 179 & 59.7 & 0.218 [0.184, 0.252] & $-$0.030 [$-$0.057, $-$0.002] & 49.7 & 42.8 & 37.4\\
Smol & 297 & 99.0 & 0.270 [0.243, 0.298] & $-$0.073 [$-$0.093, $-$0.052] & 60.3 & 43.0 & 29.0\\
Smol majority & 289 & 96.3 & 0.255 [0.227, 0.283] & $-$0.056 [$-$0.078, $-$0.034] & 57.4 & 43.1 & 30.8\\
Smol agree 2/3 & 131 & 43.7 & 0.216 [0.177, 0.256] & $-$0.022 [$-$0.055, 0.011] & 49.6 & 42.7 & 38.2\\
Smol unanimous & 7 & 2.3 & 0.208 [0.065, 0.369] & 0.043 [$-$0.108, 0.200] & 57.1 & 50.0 & 42.9\\
Cross-model & 59 & 19.7 & 0.249 [0.187, 0.316] & $-$0.060 [$-$0.111, $-$0.011] & 54.2 & 39.0 & 33.9\\
Qwen, judges disagree & 241 & 80.3 & 0.138 [0.114, 0.163] & 0.064 [0.039, 0.087] & 32.0 & 44.1 & 55.2\\
Oracle & 300 & 100.0 & 0.000 [0.000, 0.000] & 0.199 [0.186, 0.213] & 0.0 & 43.1 & 100.0\\
\bottomrule
\end{tabular*}

\captionof{table}{All policies. Cov.: share of 300 prompts acted on (\%). BMR, Rand.\ BMR, and Best in \%. ``Qwen, not unanimous'' and ``Qwen, judges disagree'' apply Qwen's original-order choice to the prompts that the respective gates reject (exploratory).}\label{tab:policies}
\vspace{1.2em}

\footnotesize
\begin{tabular*}{\textwidth}{@{\extracolsep{\fill}}lrrrrrr}
\toprule
Policy & Med. regret & BMR$_{.05}$ & Rand. BMR$_{.05}$ & Near-best & Below median & Overall regret\\
\midrule
Random & 0.167 & 37.0 & 37.0 & 42.6 & 34.5 & 0.201\\
Qwen & 0.000 & 30.7 & 37.0 & 51.7 & 28.7 & 0.178\\
Smol & 0.250 & 52.5 & 36.8 & 29.6 & 49.2 & 0.268\\
Qwen unanimous & 0.000 & 12.4 & 37.6 & 71.9 & 12.4 & 0.084\\
\bottomrule
\end{tabular*}

\captionof{table}{Sensitivity to outcome definitions (rates in \%).}\label{tab:sensitivity}
\vspace{1.2em}

\scriptsize
\begin{tabular*}{\textwidth}{@{\extracolsep{\fill}}lccccc}
\toprule
Policy & $\Delta$Stereotype & $\Delta$Miss. explicit & $\Delta$Miss. implicit & $\Delta$Quality & Overall gain\\
\midrule
Qwen & 0.023 [0.002, 0.045] & $-$0.056 [$-$0.084, $-$0.026] & $-$0.034 [$-$0.057, $-$0.009] & $-$0.031 [$-$0.052, $-$0.008] & 0.023 [0.000, 0.046]\\
Smol & $-$0.012 [$-$0.032, 0.009] & 0.110 [0.082, 0.139] & 0.030 [0.007, 0.054] & $-$0.038 [$-$0.064, $-$0.014] & $-$0.069 [$-$0.088, $-$0.050]\\
Qwen unanimous & 0.010 [$-$0.019, 0.039] & $-$0.174 [$-$0.216, $-$0.130] & $-$0.091 [$-$0.125, $-$0.056] & 0.054 [0.024, 0.082] & 0.127 [0.098, 0.155]\\
Cross-model & 0.014 [$-$0.039, 0.071] & 0.065 [0.006, 0.126] & 0.046 [$-$0.008, 0.103] & $-$0.154 [$-$0.207, $-$0.101] & $-$0.098 [$-$0.147, $-$0.050]\\
Qwen, Bonf. & [$-$0.002, 0.051] & [$-$0.090, $-$0.020] & [$-$0.062, $-$0.004] & -- & --\\
\bottomrule
\end{tabular*}

\captionof{table}{Selected minus matched random for secondary outcomes, with pointwise 95\% bootstrap intervals. For error ratings, negative values mean fewer errors; for quality and overall gain, positive values are better.}\label{tab:outcomes}
\end{table*}

\begin{table*}[p]
\centering

\footnotesize
\begin{tabular*}{\textwidth}{@{\extracolsep{\fill}}lrrlrrrr}
\toprule
Country & Rand. & Qwen & Qwen gain [95\% CI] & BMR & Smol & Smol gain & 3/3: $n$ / regret\\
\midrule
Brazil & 0.171 & 0.116 & 0.055 [0.000, 0.109] & 33.3 & 0.184 & $-$0.021 & 16 / 0.068\\
Canada & 0.207 & 0.100 & 0.107 [0.042, 0.173] & 16.7 & 0.267 & $-$0.059 & 17 / 0.010\\
Chile & 0.173 & 0.122 & 0.051 [$-$0.020, 0.119] & 33.3 & 0.189 & $-$0.016 & 16 / 0.115\\
China & 0.202 & 0.203 & $-$0.001 [$-$0.047, 0.048] & 46.7 & 0.289 & $-$0.087 & 4 / 0.104\\
Germany & 0.158 & 0.139 & 0.019 [$-$0.042, 0.080] & 36.7 & 0.264 & $-$0.106 & 13 / 0.083\\
India & 0.163 & 0.181 & $-$0.018 [$-$0.080, 0.046] & 60.0 & 0.193 & $-$0.030 & 9 / 0.083\\
Iran & 0.227 & 0.164 & 0.063 [$-$0.042, 0.160] & 26.7 & 0.306 & $-$0.079 & 14 / 0.048\\
Japan & 0.266 & 0.275 & $-$0.009 [$-$0.093, 0.073] & 46.7 & 0.400 & $-$0.134 & 9 / 0.056\\
Poland & 0.217 & 0.160 & 0.058 [$-$0.011, 0.126] & 33.3 & 0.313 & $-$0.106 & 15 / 0.106\\
South Africa & 0.206 & 0.139 & 0.067 [$-$0.003, 0.136] & 30.0 & 0.296 & $-$0.089 & 8 / 0.104\\
\bottomrule
\end{tabular*}

\captionof{table}{Country-group results (30 prompts each). Rand., Qwen, Smol: mean regret. Gains are paired with random on the same prompts. BMR in \%. Intervals are pointwise.}\label{tab:countries}
\vspace{1.2em}

\footnotesize
\begin{tabular*}{\textwidth}{@{\extracolsep{\fill}}lrrll}
\toprule
Category & $n$ & Rand. regret & Qwen regret [95\% CI] & Qwen gain [95\% CI]\\
\midrule
Dates of significance & 103 & 0.189 & 0.129 [0.094, 0.166] & 0.061 [0.025, 0.095]\\
Etiquette & 79 & 0.196 & 0.185 [0.134, 0.237] & 0.012 [$-$0.033, 0.056]\\
Family & 37 & 0.214 & 0.151 [0.091, 0.217] & 0.063 [$-$0.009, 0.136]\\
Greetings & 45 & 0.204 & 0.143 [0.087, 0.203] & 0.062 [0.009, 0.116]\\
Religion & 36 & 0.211 & 0.225 [0.148, 0.308] & $-$0.013 [$-$0.086, 0.058]\\
\bottomrule
\end{tabular*}

\captionof{table}{Category results for Qwen's original-order choices (descriptive).}\label{tab:categories}
\vspace{1.2em}

\footnotesize
\begin{tabular*}{\textwidth}{@{\extracolsep{\fill}}lrrrrrrrr}
\toprule
Judge & Calls & Valid & Parse fail & Runtime fail & Exact schema & Recovered & Non-JSON & Hours\\
\midrule
Qwen & 900 & 900 & 0 & 0 & 900 & 0 & 0 & 1.15\\
Smol & 900 & 886 & 14 & 0 & 0 & 886 & 880 & 0.27\\
\bottomrule
\end{tabular*}

\captionof{table}{Output validity over all planned calls. Hours is total recorded inference time.}\label{tab:validity}
\end{table*}

\FloatBarrier
\section{Complete Policy Results}
Table~\ref{tab:policies} lists every policy, including majority and two-of-three rules for both judges and the complements of each agreement gate. Gains are paired with exact random selection on the same prompts; BMR is compared with the random event probability on those prompts. Table~\ref{tab:sensitivity} reports alternative definitions: median regret, a 0.05 margin for below-mean selection, near-best selection, selection below the candidate median, and regret measured with overall satisfaction instead of prompt alignment. Conclusions are unchanged under these alternatives.

\section{Secondary Outcomes}
Table~\ref{tab:outcomes} reports paired differences from matched random selection for each cultural-error rating and image quality, together with the overall-satisfaction gain. For Qwen, the last row gives Bonferroni-adjusted intervals across the three cultural-error outcomes. Qwen's choices reduce missing explicit and implicit expectations under this adjustment; the stereotype increase does not survive it.

\section{Country and Category Results}
Table~\ref{tab:countries} expands Figure~\ref{fig:country}. The last column reports how many prompts per country pass the unanimity gate and their regret; the gate represents all countries, but acceptance ranges widely across groups. Table~\ref{tab:categories} reports the declared category analysis.

\section{Ablation Details}
Table~\ref{tab:ablations} lists every condition summarized in Section~\ref{sec:ablations}, and Table~\ref{tab:scale} the scale extension of Section~\ref{sec:scale}. Taken alone, the added fourth rotation is not balanced across generators, so only the combined four-order results are reported.
\begin{table*}[t]
\centering
\scriptsize
\begin{tabular*}{\textwidth}{@{\extracolsep{\fill}}lrrrll}
\toprule
Condition & First slot (\%) & Uniform (\%) & Choice changes (\%) & Gain [95\% CI] & Unanimous: coverage, gain [95\% CI]\\
\midrule
Main (3 orders) & 48.8 & 27.8 & 59.7 & 0.039 [0.016, 0.062] & 40.3\%, 0.141 [0.110, 0.173]\\
No ranking (choice-only format) & 44.8 & 27.8 & 54.0 & 0.047 [0.024, 0.069] & 46.0\%, 0.123 [0.094, 0.153]\\
Opaque codes (no letters) & 51.8 & 27.8 & 62.7 & 0.021 [$-$0.001, 0.044] & 37.3\%, 0.124 [0.087, 0.160]\\
Four orders (4-image pools) & 52.5 & 25.0 & 72.5 & 0.028 [$-$0.000, 0.056] & 27.5\%, 0.127 [0.073, 0.178]\\
\bottomrule
\end{tabular*}

\caption{Post-hoc ablations (Qwen, 300 prompts; four orders use the 200 four-image pools). Choice changes: share of prompts whose returned image differs across orders. Unanimous: prompts where all orders return the same image.}
\label{tab:ablations}
\end{table*}

\begin{table*}[t]
\centering
\resizebox{\textwidth}{!}{\begin{tabular}{lrrlrrlll}
\toprule
Judge & First slot (\%) & Choice changes (\%) & Gain [95\% CI] & Regret & BMR (\%) & Unanimous: cov.\ (\%), gain & Rejected: gain [95\% CI] & $\Delta$Stereo.\ [95\% CI]\\
\midrule
Qwen3-VL-4B (main, bf16) & 48.8 & 59.7 & 0.039 [0.016, 0.062] & 0.160 & 36.3 & 40.3, 0.141 & $-$0.030 [$-$0.057, $-$0.002] & 0.023 [0.002, 0.045]\\
Qwen3-VL-4B (8-bit control) & 48.2 & 57.3 & 0.044 [0.022, 0.066] & 0.155 & 36.0 & 42.7, 0.135 & $-$0.023 [$-$0.051, 0.004] & 0.021 [0.001, 0.043]\\
Qwen3-VL-8B (8-bit) & 27.3 & 43.0 & 0.097 [0.077, 0.117] & 0.102 & 24.0 & 57.0, 0.118 & 0.068 [0.036, 0.099] & 0.016 [$-$0.004, 0.036]\\
\bottomrule
\end{tabular}
}
\caption{Scale extension (300 prompts, three orders). The 8-bit runs use Apple MLX with the same inputs, decoding, and parser as the main run; the 8-bit 4B run controls for quantization and agrees with the main 4B run on most calls. Rejected: prompts on which the three orders disagree.}
\label{tab:scale}
\end{table*}

\section{Content-Only Baseline and Annotation Robustness}
Table~\ref{tab:baselines} compares the CLIP selector with the judges; Table~\ref{tab:robustness} repeats the Qwen analysis on prompts split by annotator agreement and on prompts with a unique best candidate. Annotator disagreement is the mean, over candidates with at least two ratings, of the variance of individual alignment ratings; the split is at its median.

\begin{table*}[t]
\centering
% CLIP openai/clip-vit-large-patch14@32bd64288804d66eefd0ccbe215aa642df71cc41; agreement Qwen 49.0\%, Smol 13.1\%, chance 27.8\%; paired CLIP$-$Qwen gain 0.030 [0.005, 0.055].

\footnotesize
\begin{tabular*}{\textwidth}{@{\extracolsep{\fill}}lrllrrrl}
\toprule
Selector & $n$ & Regret [95\% CI] & Gain [95\% CI] & BMR & Rand. BMR & Best & $\Delta$Stereotype [95\% CI]\\
\midrule
Random & 300 & 0.199 [0.186, 0.213] & 0.000 [0.000, 0.000] & 43.1 & 43.1 & 41.9 & --\\
CLIP & 300 & 0.130 [0.109, 0.153] & 0.069 [0.047, 0.090] & 28.0 & 43.1 & 55.7 & 0.026 [0.006, 0.046]\\
Qwen & 300 & 0.160 [0.136, 0.184] & 0.039 [0.016, 0.062] & 36.3 & 43.1 & 51.0 & 0.023 [0.002, 0.045]\\
Smol & 297 & 0.270 [0.243, 0.298] & $-$0.073 [$-$0.093, $-$0.052] & 60.3 & 43.0 & 29.0 & $-$0.012 [$-$0.032, 0.009]\\
Oracle & 300 & 0.000 [0.000, 0.000] & 0.199 [0.186, 0.213] & 0.0 & 43.1 & 100.0 & $-$0.020 [$-$0.034, $-$0.006]\\
\bottomrule
\end{tabular*}

\vspace{4pt}

\footnotesize
\begin{tabular*}{\textwidth}{@{\extracolsep{\fill}}llll}
\toprule
Selector & $\Delta$Stereotype & $\Delta$Miss. explicit & $\Delta$Miss. implicit\\
\midrule
CLIP & 0.026 [0.006, 0.046] & $-$0.113 [$-$0.141, $-$0.083] & 0.003 [$-$0.022, 0.028]\\
Qwen & 0.023 [0.002, 0.045] & $-$0.056 [$-$0.084, $-$0.026] & $-$0.034 [$-$0.057, $-$0.009]\\
\bottomrule
\end{tabular*}

\vspace{4pt}

\footnotesize
\begin{tabular*}{\textwidth}{@{\extracolsep{\fill}}lrlr}
\toprule
Prompts & $n$ & Qwen gain [95\% CI] & Qwen regret\\
\midrule
Qwen agrees with CLIP & 147 & 0.099 [0.067, 0.130] & 0.111\\
Qwen disagrees with CLIP & 153 & $-$0.018 [$-$0.047, 0.011] & 0.206\\
\bottomrule
\end{tabular*}

\caption{Content-only CLIP selector (argmax prompt--image similarity, position-invariant by construction) versus the judges, original order. BMR, Rand.\ BMR, and Best in \%. $\Delta$Stereotype is selected minus matched random.}
\label{tab:baselines}
\end{table*}

\begin{table*}[t]
\centering
% Disagreement median 0.0417; high consensus n=138, low n=162; unique best n=181.

\scriptsize
\begin{tabular*}{\textwidth}{@{\extracolsep{\fill}}llrrllrrl}
\toprule
Subset & Policy & $n$ & Rand. regret & Regret [95\% CI] & Gain [95\% CI] & BMR & Rand. BMR & $\Delta$Stereotype [95\% CI]\\
\midrule
High consensus & Qwen & 138 & 0.185 & 0.138 [0.106, 0.172] & 0.047 [0.014, 0.079] & 31.9 & 40.3 & 0.033 [0.004, 0.063]\\
High consensus & Qwen unanimous & 60 & 0.217 & 0.072 [0.037, 0.111] & 0.146 [0.100, 0.192] & 15.0 & 42.1 & 0.009 [$-$0.029, 0.049]\\
Low consensus & Qwen & 162 & 0.211 & 0.178 [0.146, 0.212] & 0.033 [0.002, 0.062] & 40.1 & 45.4 & 0.015 [$-$0.014, 0.045]\\
Low consensus & Qwen unanimous & 61 & 0.213 & 0.076 [0.045, 0.111] & 0.137 [0.095, 0.178] & 18.0 & 44.9 & 0.012 [$-$0.029, 0.054]\\
Unique best & Qwen & 181 & 0.250 & 0.195 [0.163, 0.229] & 0.056 [0.025, 0.087] & 38.7 & 49.2 & 0.014 [$-$0.012, 0.040]\\
Unique best & Qwen unanimous & 80 & 0.260 & 0.092 [0.060, 0.127] & 0.169 [0.127, 0.210] & 18.8 & 48.8 & $-$0.008 [$-$0.040, 0.024]\\
\bottomrule
\end{tabular*}

\vspace{4pt}

\footnotesize
\begin{tabular*}{\textwidth}{@{\extracolsep{\fill}}lrrrrrrr}
\toprule
Category & Qwen calls & Qwen A (\%) & Unif. (\%) & Qwen flip (\%) & Smol A (\%) & Unif. (\%) & Smol flip (\%)\\
\midrule
Dates of significance & 309 & 48.2 & 28.6 & 56.3 & 29.9 & 28.6 & 98.0\\
Etiquette & 237 & 49.4 & 26.8 & 63.3 & 45.1 & 26.8 & 97.4\\
Family & 111 & 44.1 & 28.6 & 51.4 & 50.9 & 28.6 & 97.2\\
Greetings & 135 & 51.9 & 27.2 & 66.7 & 63.6 & 27.2 & 97.6\\
Religion & 108 & 50.0 & 27.3 & 61.1 & 41.9 & 27.3 & 97.1\\
All & 900 & 48.8 & 27.8 & 59.7 & 43.0 & 27.8 & 97.6\\
\bottomrule
\end{tabular*}

\caption{Robustness to annotation consensus (exploratory). Rand.\ regret is random choice on the same prompts.}
\label{tab:robustness}
\end{table*}

\FloatBarrier
\section{Additional Analyses}
\label{app:extras}
\begin{figure}[t]
\centering
\includegraphics[width=\columnwidth]{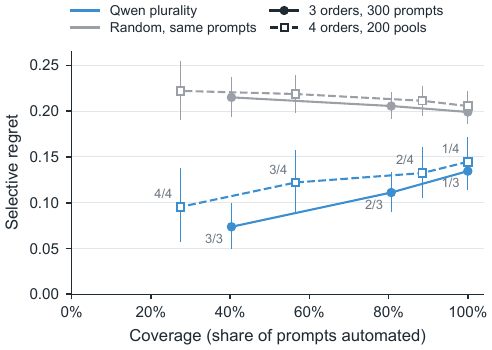}
\caption{Risk--coverage operating points for agreement gates with three and four orders, next to matched random choice (95\% intervals).}
\label{fig:risk-coverage}
\end{figure}

This section reports analyses requested in internal review: the configuration and backbone sensitivity of the CLIP baseline (Table~\ref{tab:extras-clip}), the cost of order checks (Table~\ref{tab:extras-latency}), the relation between order sensitivity and pool difficulty (Table~\ref{tab:extras-difficulty}), a decomposition of the stereotype result (Tables~\ref{tab:extras-stereo}--\ref{tab:extras-genslot}), cultural-error profiles of aggregated policies (Table~\ref{tab:extras-profiles}), and risk--coverage operating points with three and four orders (Figure~\ref{fig:risk-coverage}, Table~\ref{tab:extras-rc}). All are post hoc and exploratory.

\begin{table*}[p]
\centering
% CLIP backbones agree on 58.7\% of prompts; B/32$-$L/14 gain $-$0.006 [$-$0.028, 0.018].

\footnotesize
\begin{tabular*}{\textwidth}{@{\extracolsep{\fill}}lrllrrl}
\toprule
Selector & $n$ & Regret [95\% CI] & Gain [95\% CI] & BMR & Rand. BMR & $\Delta$Stereotype [95\% CI]\\
\midrule
Random & 300 & 0.199 [0.186, 0.213] & 0.000 [0.000, 0.000] & 43.1 & 43.1 & --\\
CLIP ViT-L/14 & 300 & 0.130 [0.109, 0.153] & 0.069 [0.047, 0.090] & 28.0 & 43.1 & 0.026 [0.006, 0.046]\\
CLIP ViT-B/32 & 300 & 0.136 [0.114, 0.159] & 0.063 [0.042, 0.085] & 32.0 & 43.1 & 0.003 [$-$0.016, 0.023]\\
Qwen & 300 & 0.160 [0.136, 0.184] & 0.039 [0.016, 0.062] & 36.3 & 43.1 & 0.023 [0.002, 0.045]\\
Oracle & 300 & 0.000 [0.000, 0.000] & 0.199 [0.186, 0.213] & 0.0 & 43.1 & $-$0.020 [$-$0.034, $-$0.006]\\
\bottomrule
\end{tabular*}

\captionof{table}{CLIP baseline with two backbones (original order). Both embed the raw prompt and each image with default CLIP preprocessing and return the most similar image. BMR in \%.}\label{tab:extras-clip}
\vspace{1em}

\footnotesize
\begin{tabular*}{\textwidth}{@{\extracolsep{\fill}}lrrrrrrrr}
\toprule
Judge & Calls & Med. s/call & Mean s/call & 1-order s/prompt & 3-order s/prompt & Ratio & 1-order h & 3-order h\\
\midrule
Qwen & 900 & 3.5 & 4.6 & 5.4 & 13.8 & 2.57 & 0.45 & 1.15\\
Smol & 900 & 1.0 & 1.1 & 1.2 & 3.2 & 2.80 & 0.10 & 0.27\\
\bottomrule
\end{tabular*}

\captionof{table}{Inference cost on Apple Silicon (MPS, bfloat16, greedy decoding). Per-prompt times compare a single order with three orders.}\label{tab:extras-latency}
\vspace{1em}
% logistic log-odds per SD: range $-$0.241 [$-$0.498, 0.045]; near_ties 0.065 [$-$0.196, 0.360]; size 0.251 [0.009, 0.504]

\footnotesize
\begin{tabular*}{\textwidth}{@{\extracolsep{\fill}}lrrrl}
\toprule
Stratum & Range & $n$ & Flip (\%) & 95\% CI\\
\midrule
Low & 0.00--0.33 & 148 & 62.8 & [54.7, 70.3]\\
Mid & 0.38--0.50 & 83 & 59.0 & [48.2, 69.9]\\
High & 0.58--1.00 & 69 & 53.6 & [42.0, 65.2]\\
1 near-tie & -- & 177 & 55.9 & [48.6, 63.3]\\
2 near-ties & -- & 87 & 65.5 & [55.2, 74.7]\\
3 near-ties & -- & 32 & 59.4 & [43.8, 75.0]\\
4 near-ties & -- & 4 & 100.0 & [100.0, 100.0]\\
\bottomrule
\end{tabular*}

\captionof{table}{Share of prompts on which Qwen's choice changes across the three orders, by tertile of the pool's rating range (best minus worst candidate).}\label{tab:extras-difficulty}
\vspace{1em}

\footnotesize
\begin{tabular*}{\textwidth}{@{\extracolsep{\fill}}lrlll}
\toprule
Policy & $n$ & $\Delta$Stereotype & $\Delta$Miss. explicit & $\Delta$Miss. implicit\\
\midrule
Qwen & 300 & 0.023 [0.002, 0.045] & $-$0.056 [$-$0.084, $-$0.026] & $-$0.034 [$-$0.057, $-$0.009]\\
Qwen majority & 300 & 0.007 [$-$0.013, 0.027] & $-$0.092 [$-$0.120, $-$0.064] & $-$0.045 [$-$0.068, $-$0.021]\\
Qwen unanimous & 121 & 0.010 [$-$0.019, 0.039] & $-$0.174 [$-$0.216, $-$0.130] & $-$0.091 [$-$0.125, $-$0.056]\\
Qwen, not unanimous & 179 & 0.032 [0.003, 0.063] & 0.025 [$-$0.010, 0.059] & 0.005 [$-$0.025, 0.037]\\
\bottomrule
\end{tabular*}

\captionof{table}{Cultural-error differences from matched random choice for Qwen's direct, majority-vote, and unanimous policies, and for the prompts unanimity rejects (95\% CI).}\label{tab:extras-profiles}
\end{table*}

\begin{table*}[p]
\centering

\scriptsize
\begin{tabular*}{\textwidth}{@{\extracolsep{\fill}}lrrrrlr}
\toprule
Chosen generator & Uniform (\%) & Qwen (\%) & Stereo. & Within-pool & $\Delta$Stereo. when chosen [95\% CI] & Contribution\\
\midrule
Flux.1-Dev & 27.3 & 3.3 & 0.074 & $-$0.051 & $-$0.089 [$-$0.164, $-$0.028] & $-$0.003\\
GPT-Image & 27.2 & 41.0 & 0.114 & $-$0.009 & 0.006 [$-$0.022, 0.035] & 0.003\\
Imagen3 & 17.9 & 14.3 & 0.146 & 0.026 & 0.042 [$-$0.021, 0.111] & 0.006\\
SD-3.5-Large & 27.6 & 41.3 & 0.165 & 0.041 & 0.042 [0.009, 0.077] & 0.018\\
All & 100.0 & 100.0 & -- & -- & 0.023 [0.002, 0.045] & 0.023\\
\bottomrule
\end{tabular*}

\captionof{table}{Stereotype decomposition by the generator of Qwen's chosen image: choice share against uniform, mean stereotype rating of the generator's images, and the stereotype difference from random when that generator is chosen.}\label{tab:extras-stereo}
\vspace{1em}

\footnotesize
\begin{tabular*}{\textwidth}{@{\extracolsep{\fill}}lrllr}
\toprule
Prompts & $n$ & Qwen $\Delta$Stereo. [95\% CI] & Oracle $\Delta$Stereo. [95\% CI] & Qwen regret\\
\midrule
Alignment-best tied & 119 & 0.038 [0.004, 0.073] & $-$0.008 [$-$0.024, 0.007] & 0.107\\
Unique alignment-best & 181 & 0.014 [$-$0.012, 0.040] & $-$0.028 [$-$0.048, $-$0.008] & 0.195\\
\bottomrule
\end{tabular*}

\captionof{table}{Stereotype difference on prompts where several candidates tie for the best alignment rating versus prompts with a unique best candidate.}\label{tab:extras-ties}
\vspace{1em}

\footnotesize
\begin{tabular*}{\textwidth}{@{\extracolsep{\fill}}lrrrrr}
\toprule
Generator & A (\%) & B (\%) & C (\%) & D (\%) & All (\%)\\
\midrule
Flux.1-Dev & 22.0 (296) & 3.0 (296) & 1.3 (158) & 0.0 (138) & 8.6\\
GPT-Image & 86.7 (233) & 61.4 (295) & 41.0 (295) & 27.4 (62) & 58.9\\
Imagen3 & 65.8 (73) & 15.4 (149) & 16.6 (211) & 19.5 (200) & 22.9\\
SD-3.5-Large & 41.6 (298) & 6.9 (160) & 6.8 (236) & 3.5 (200) & 17.7\\
\bottomrule
\end{tabular*}

\captionof{table}{Qwen's choice rate for each generator by presentation slot, pooled over the three orders (number of presentations in parentheses).}\label{tab:extras-genslot}
\vspace{1em}

\footnotesize
\begin{tabular*}{\textwidth}{@{\extracolsep{\fill}}lrrrlll}
\toprule
Protocol & $q\ge$ & $n$ & Cov. (\%) & Regret [95\% CI] & Rand. regret [95\% CI] & Difference [95\% CI]\\
\midrule
3 orders, all prompts & 1/3 & 300 & 100.0 & 0.135 [0.114, 0.156] & 0.199 [0.186, 0.213] & $-$0.065 [$-$0.086, $-$0.043]\\
3 orders, all prompts & 2/3 & 242 & 80.7 & 0.111 [0.090, 0.133] & 0.206 [0.192, 0.220] & $-$0.094 [$-$0.118, $-$0.071]\\
3 orders, all prompts & 3/3 & 121 & 40.3 & 0.074 [0.050, 0.100] & 0.215 [0.193, 0.237] & $-$0.141 [$-$0.173, $-$0.110]\\
3 orders, 4-image pools & 1/3 & 200 & 100.0 & 0.151 [0.125, 0.179] & 0.206 [0.190, 0.222] & $-$0.055 [$-$0.081, $-$0.028]\\
3 orders, 4-image pools & 2/3 & 159 & 79.5 & 0.124 [0.097, 0.152] & 0.210 [0.193, 0.228] & $-$0.086 [$-$0.114, $-$0.057]\\
3 orders, 4-image pools & 3/3 & 73 & 36.5 & 0.080 [0.049, 0.114] & 0.219 [0.194, 0.245] & $-$0.139 [$-$0.179, $-$0.097]\\
4 orders, 4-image pools & 1/4 & 200 & 100.0 & 0.145 [0.118, 0.172] & 0.206 [0.190, 0.222] & $-$0.061 [$-$0.087, $-$0.034]\\
4 orders, 4-image pools & 2/4 & 177 & 88.5 & 0.132 [0.105, 0.161] & 0.211 [0.195, 0.228] & $-$0.079 [$-$0.106, $-$0.050]\\
4 orders, 4-image pools & 3/4 & 113 & 56.5 & 0.122 [0.090, 0.157] & 0.219 [0.198, 0.240] & $-$0.097 [$-$0.131, $-$0.061]\\
4 orders, 4-image pools & 4/4 & 55 & 27.5 & 0.095 [0.058, 0.138] & 0.222 [0.191, 0.255] & $-$0.127 [$-$0.178, $-$0.073]\\
\bottomrule
\end{tabular*}

\captionof{table}{Risk--coverage operating points: selective regret on kept prompts versus matched random choice, for agreement thresholds with three orders (all prompts and four-image pools) and four orders (four-image pools).}\label{tab:extras-rc}
\end{table*}

\section{Presentation Position}
Table~\ref{tab:position} separates position bias by pool size. Qwen favors slot A in both pool sizes. Smol favors slot B for three-image pools and slots A and B for four-image pools, and rarely chooses later slots.

\begin{table*}[t]
\centering

\footnotesize
\begin{tabular*}{\textwidth}{@{\extracolsep{\fill}}lrrrrrrr}
\toprule
Judge & Pool & Calls & A (\%) & B (\%) & C (\%) & D (\%) & Uniform (\%)\\
\midrule
Qwen & 3 & 300 & 54.3 & 25.3 & 20.3 & -- & 33.3\\
Qwen & 4 & 600 & 46.0 & 24.7 & 18.8 & 10.5 & 25.0\\
Smol & 3 & 294 & 19.0 & 71.8 & 9.2 & -- & 33.3\\
Smol & 4 & 592 & 54.9 & 39.7 & 1.7 & 3.7 & 25.0\\
\bottomrule
\end{tabular*}

\caption{Choice share by presentation slot and pool size (\%), over all recorded orders.}
\label{tab:position}
\end{table*}

\section{Output Validity}
Table~\ref{tab:validity} summarizes parsing outcomes. Qwen satisfied the strict two-field schema on every call. Smol rarely produced valid JSON; its choices were recovered by the predeclared single-label normalization described below, and every recovery is flagged in the logs.

\section{Design Choices and Anticipated Questions}
\paragraph{Why prompt alignment as the primary outcome?} It is the rating most directly tied to whether a returned image satisfies the request, and it is available for every candidate. Cultural-error ratings are reported separately rather than folded into a composite, because the results show they can move in different directions.
\paragraph{Why not debias position directly?} Order averaging or position-balanced prompting would change the frozen policy after observing results. We report the bias and leave debiased selectors to future work, where they can be evaluated with the same audit.
\paragraph{What happens to abstained requests?} We do not model a human fallback. Coverage is reported alongside every selective estimate so that the cost of abstention remains visible.

\section{Interactive Audit Browser}
\label{app:browser}
\begin{figure*}[p]
\centering
\includegraphics[width=0.86\textwidth]{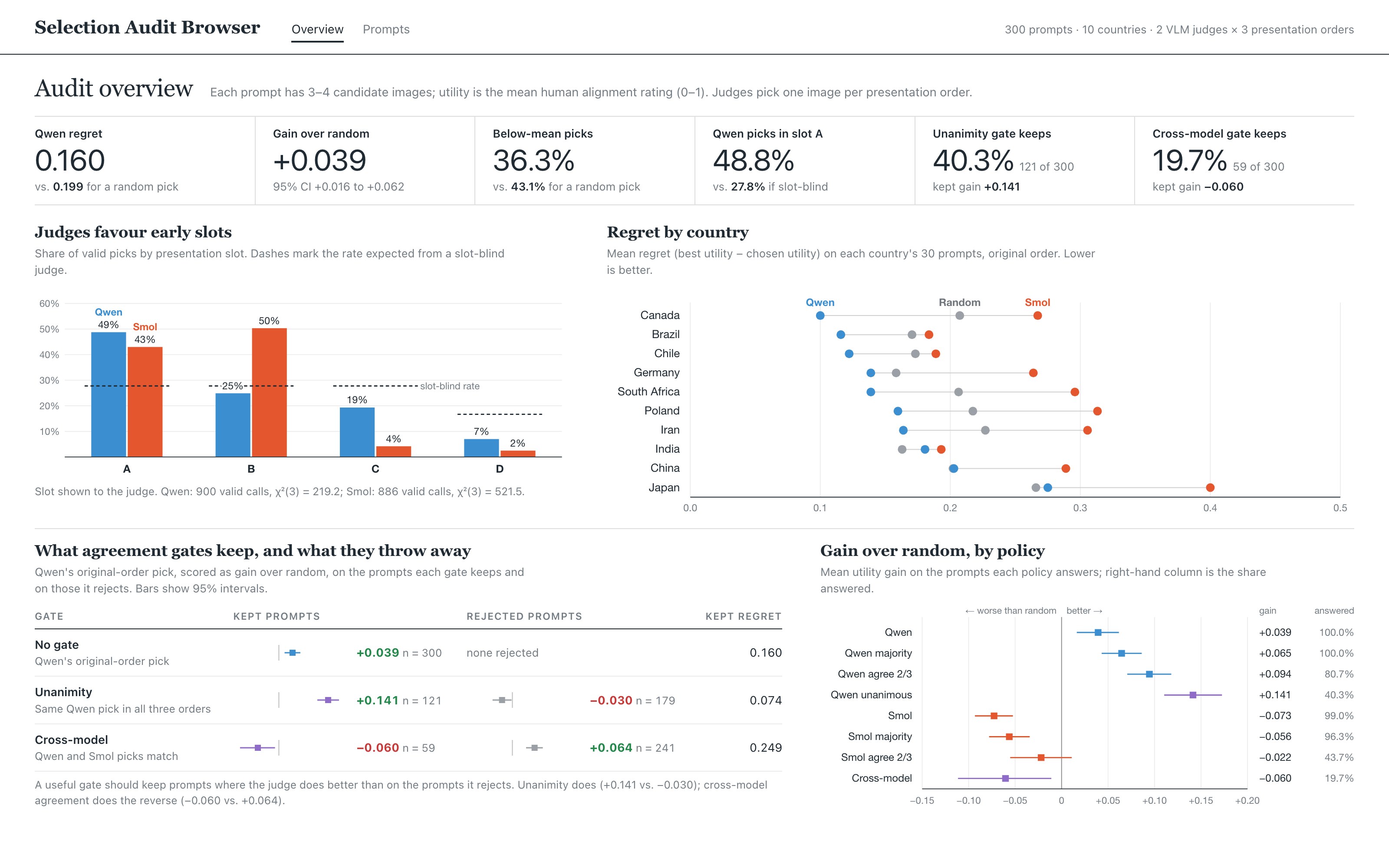}
\caption{Audit browser, overview. Headline estimates, presentation-slot preference, country-group regret, kept-versus-rejected gains for each gate, and gain over random for every policy, all computed from the recorded outputs.}
\label{fig:browser-overview}
\vspace{1.4em}
\includegraphics[width=0.86\textwidth]{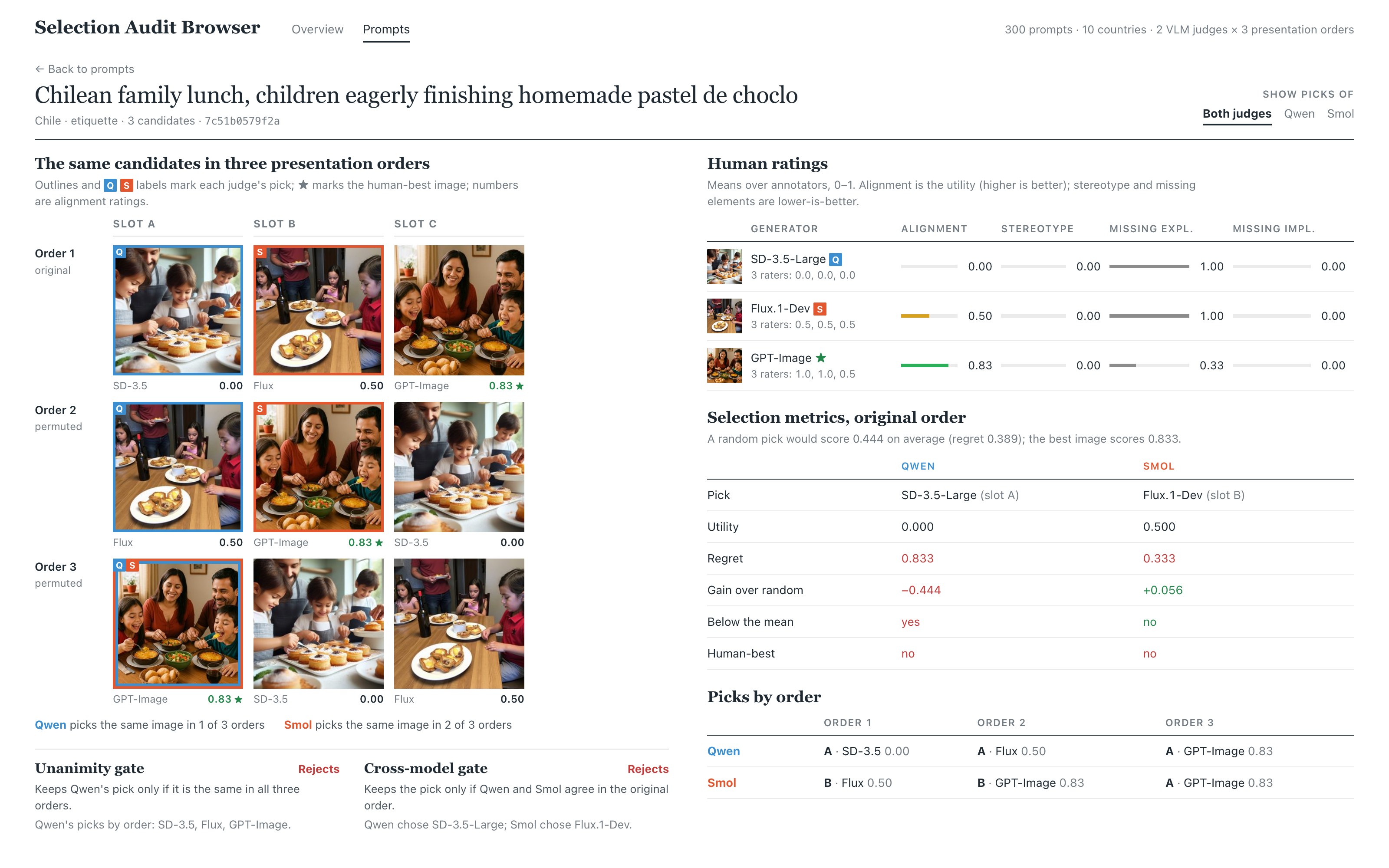}
\caption{Audit browser, detail view for the prompt in Figure~\ref{fig:hero}. Left: the three presentation orders with each judge's choice (Q, S) and the best-rated image (check). Right: per-candidate human ratings, selection metrics for the original order, and choices by order. Images from CulturalFrames \citep{culturalframes}.}
\label{fig:browser-detail}
\end{figure*}

\begin{figure*}[p]
\centering
\includegraphics[width=0.86\textwidth]{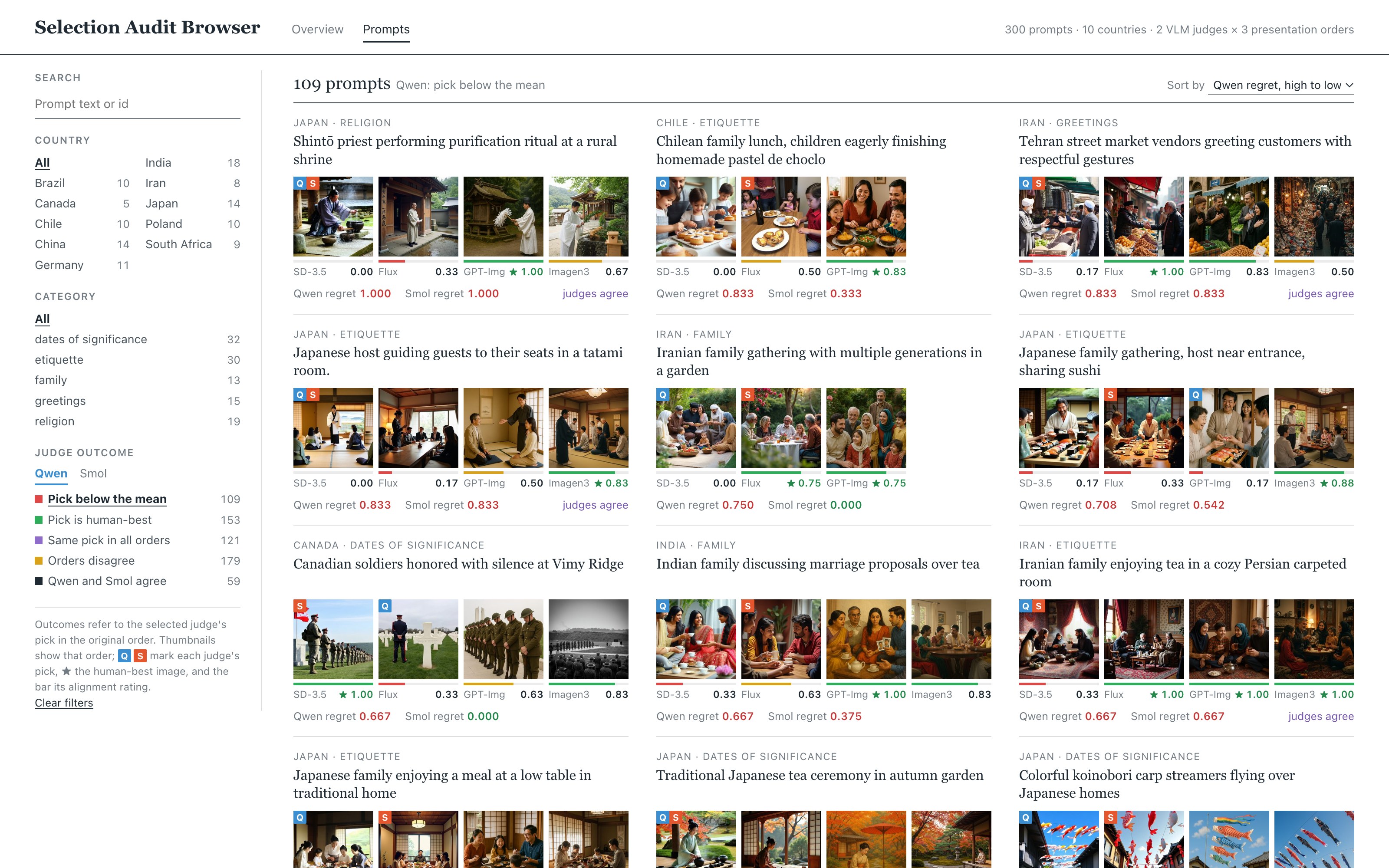}
\caption{Audit browser, prompt explorer filtered to Qwen's below-mean choices and sorted by regret. Each entry shows the pool with both judges' original-order choices (Q, S), the best-rated image (star), and the ratings.}
\label{fig:browser-below}
\vspace{1.4em}
\includegraphics[width=0.86\textwidth]{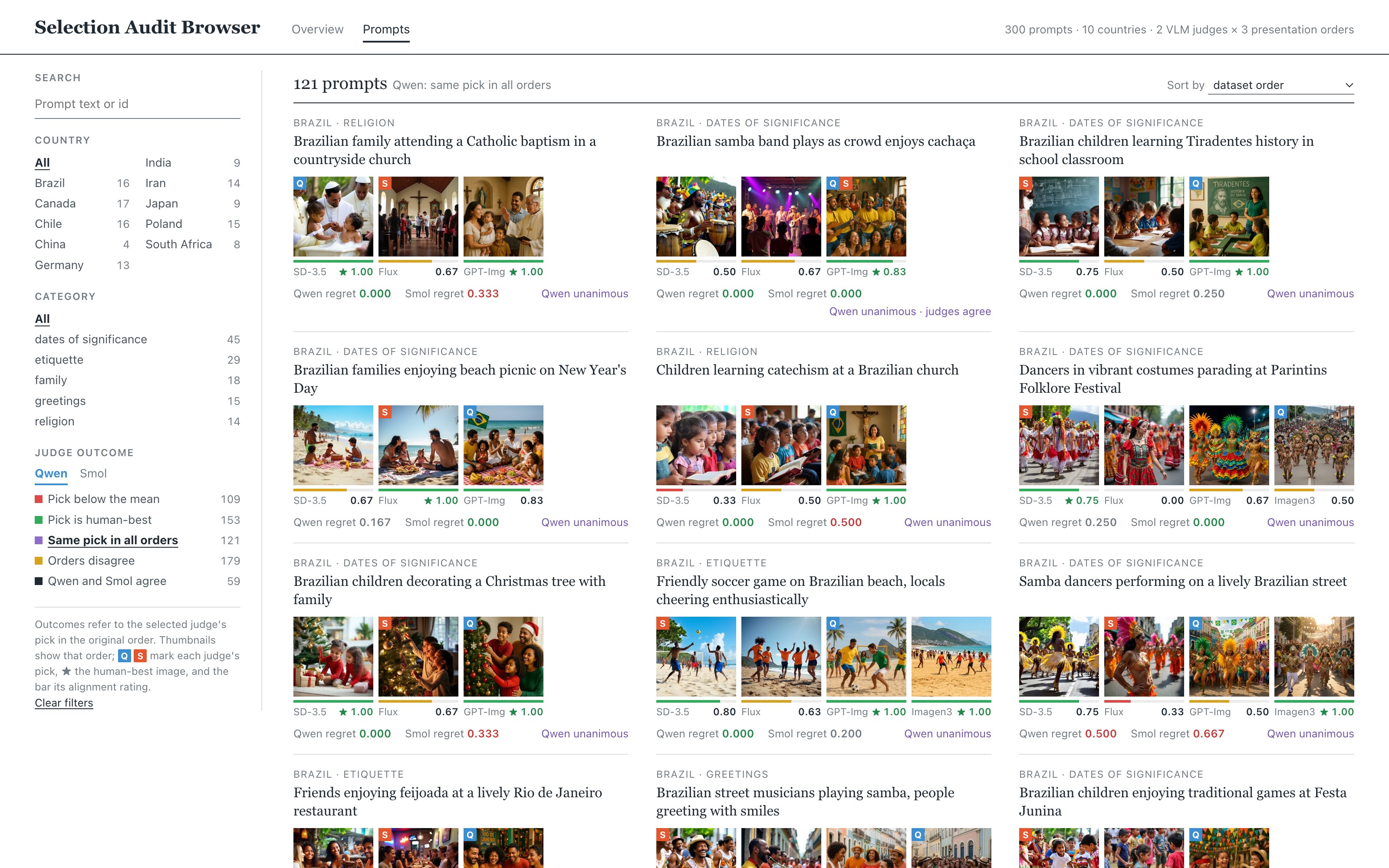}
\caption{Audit browser, prompt explorer filtered to prompts on which Qwen picks the same image in all three orders (the unanimity gate keeps these).}
\label{fig:browser-unanimous}
\end{figure*}

\begin{figure*}[p]
\centering
\includegraphics[width=0.86\textwidth]{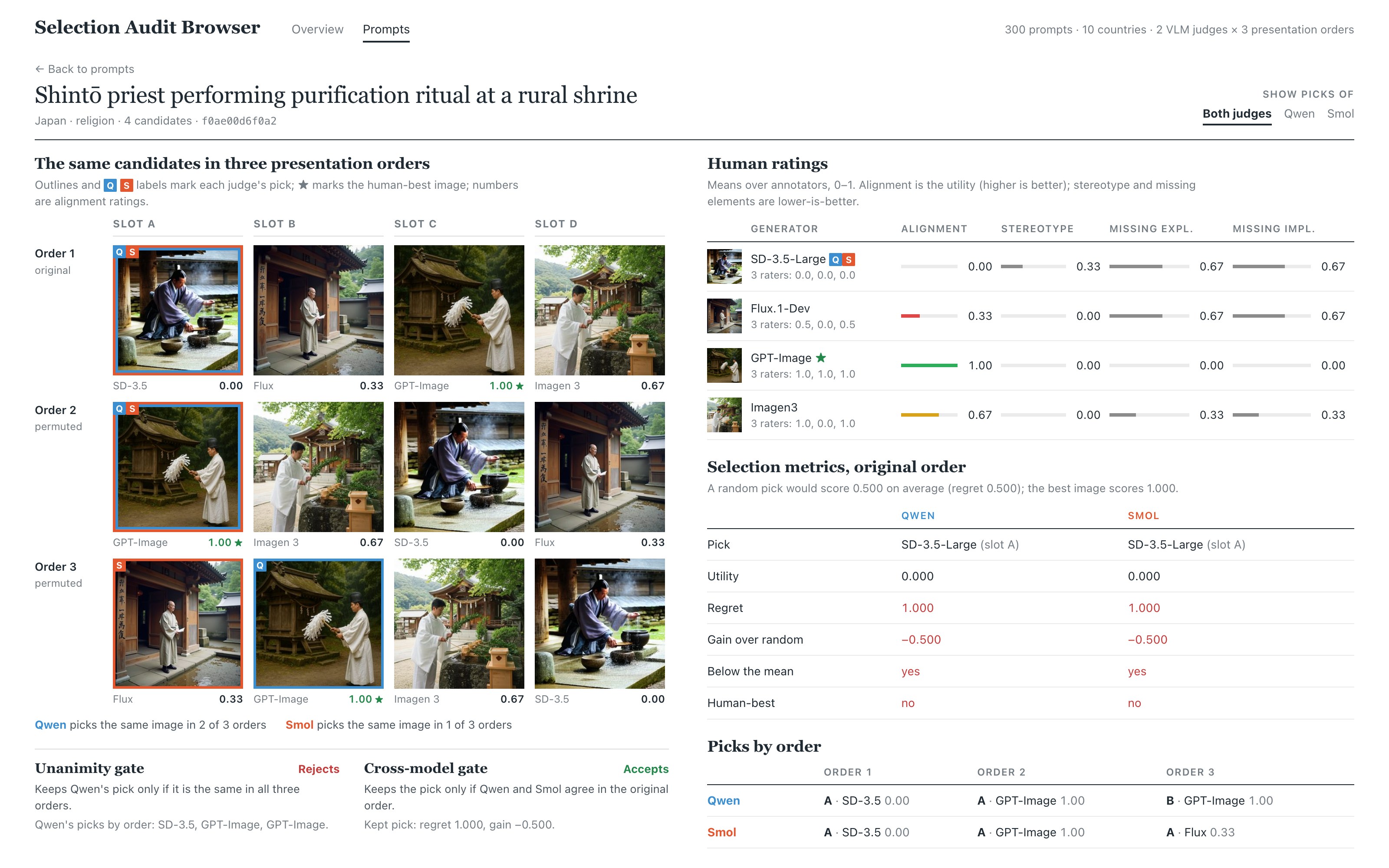}
\caption{Audit browser, detail view for the four-image pool with the largest Qwen regret (selected by rule). Both judges return the lowest-rated image, so the cross-model gate would act on this prompt.}
\label{fig:browser-detail2}
\vspace{1.4em}
\includegraphics[width=0.86\textwidth]{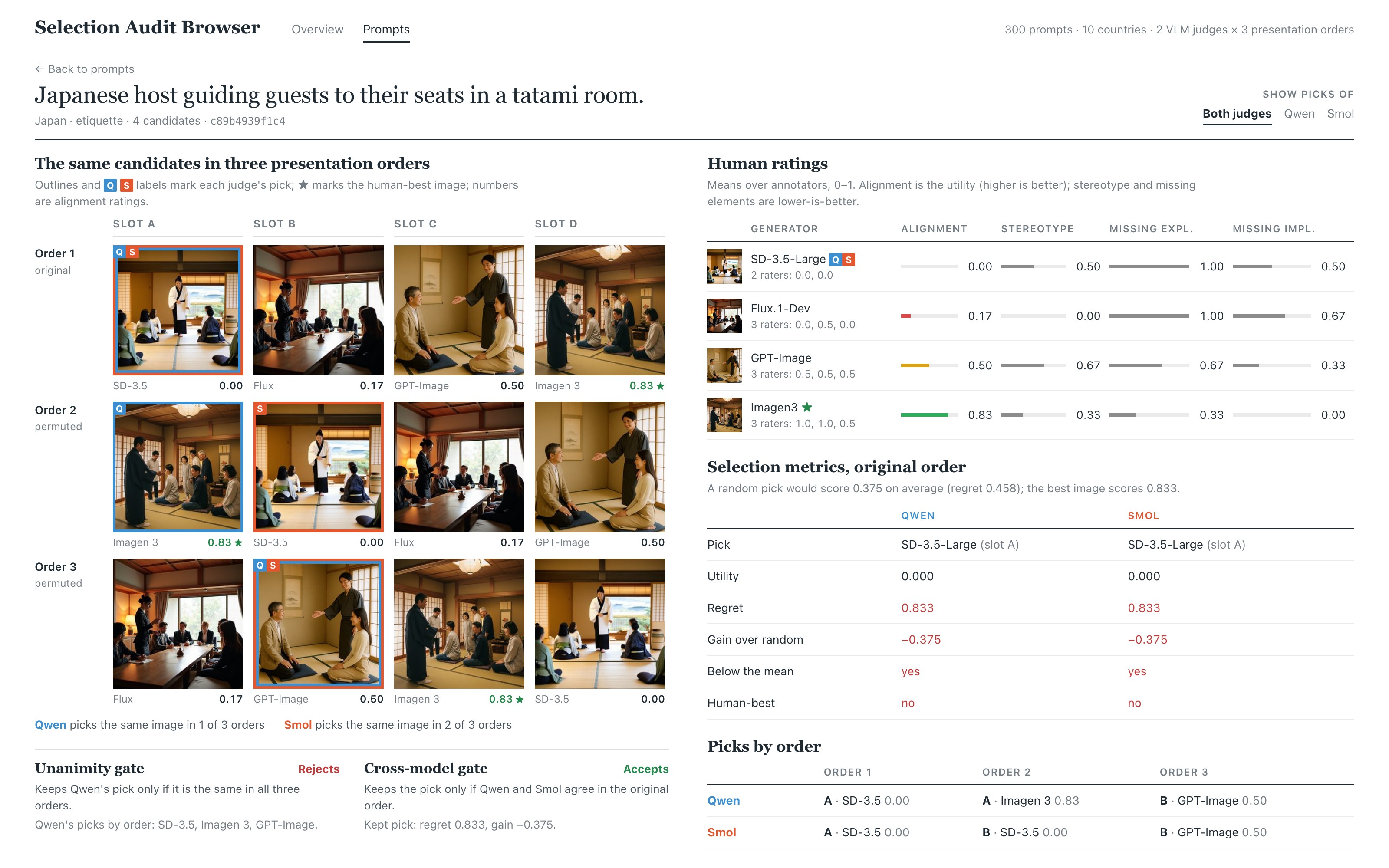}
\caption{Audit browser, detail view for a prompt where the two judges agree on a poorly rated image (next-largest regret among cross-model agreements).}
\label{fig:browser-detail3}
\end{figure*}

To make every decision inspectable, the artifact includes an offline browser built only from the recorded judge logs and released ratings (Figures~\ref{fig:browser-overview}--\ref{fig:browser-detail3}). The overview reproduces the headline estimates, slot preference, country regret, and the kept-versus-rejected gains of each gate. The prompt explorer filters pools by country, category, and outcome (for example, below-mean choices or prompts where the orders disagree). The detail view shows each pool under all three orders with both judges' choices, the per-candidate ratings, and whether each gate would act. The browser runs locally, needs no network access, and does not redistribute source images.

\FloatBarrier
\section{Estimation Details}
\label{app:stats}
We use 10,000 nonparametric prompt-level bootstrap resamples with seed 20261002 for means and paired gains. Each prompt's candidates and decisions travel together; permutation calls are never independently resampled. Binary deterministic-policy BMR uses Wilson intervals. Random-policy BMR averages fractional within-prompt probabilities and uses a bootstrap interval. Country results have subgroup bootstrap intervals, and the worst-group statistic is recomputed within resamples rather than selecting the upper endpoint of a single country's interval.

The pooled prompt bootstrap approximates sampling from the balanced source mixture; it does not hold country counts fixed in every replicate. Worst-group resampling is stratified within represented countries, but its percentile interval is descriptive and need not be well calibrated when group means are tied or nearly tied. An interval for the group with the largest observed mean is not simultaneous or selection-adjusted.

All intervals condition on the released annotation means. They do not account for annotation uncertainty or recruitment bias. Estimates after abstention weight accepted prompts equally, so country weights can differ from the balanced source split. We report group coverage and compare random selection on the same accepted prompts. The original endpoints, acceptance thresholds, and sensitive-set definitions remain fixed throughout the main run.

Intervals for the secondary cultural-rating outcomes are pointwise, without adjustment for multiple comparisons. We interpret them as descriptive evidence about the released ratings, not confirmatory evidence of cultural harm across populations.

\paragraph{Random-choice below-mean probability.} For random choice the below-mean event is a probability, $\frac{1}{m_p}\sum_i \mathbf{1}\{h_{pi}<\bar h_p\}$, averaged over the prompts in $A_\pi$. Replacing it with the indicator of the expected utility would make it identically zero. A sensitivity margin $\delta=0.05$ replaces $\bar h_p$ by $\bar h_p-\delta$ (Table~\ref{tab:sensitivity}).
\paragraph{Order-consistency entropy.} We also record the normalized entropy of the three choices, $H_p=-\sum_i f_{pi}\log f_{pi}/\log m_p$, where $f_{pi}$ is the share of orders choosing $x_{pi}$. With three calls it can reach one for three candidates but at most $\log 3/\log 4$ for four, so it is not comparable across pool sizes; it plays no role in the gates.
\paragraph{Worst-group regret.} For country group $g$, $R_g$ is the mean regret over acted-on prompts in $g$; worst-group regret is $\max_g R_g$ and disparity is $\max_g R_g-\min_g R_g$. Groups with no accepted prompts are reported as zero coverage, not zero risk.

\section{Judge Instruction}
The shared instruction reads:
\begin{quote}
You are evaluating candidate images for a generative image system. Select the ONE image that should be returned to the user.

Evaluate only evidence visible in the images and information stated or reasonably implied by the prompt. Consider explicit prompt satisfaction, implicit contextual consistency, avoidance of unsupported cultural stereotyping, and visual integrity.

Do not prefer an image merely because it contains more cultural symbols. Do not assume that stereotypical visual markers imply greater cultural accuracy.

The candidate labels are supplied next to each image. Return JSON only, with choice equal to the first element of ranking and ranking containing every candidate label exactly once. Do not explain your decision.
\end{quote}
The user prompt follows this instruction, then each label and image are interleaved. A final text segment states the allowed labels and requires the two JSON keys, without specifying a candidate as an example answer. No human annotation, generator name, country field, or category field is included.

\section{Invalid Output and Reproducibility}
Qwen's strict parser accepts exactly the two requested fields, a complete permutation of supplied labels, and a choice equal to the first ranking entry. It rejects explanatory wrappers, repeated labels, incomplete rankings, and inconsistent top choices. Smol requests exactly one \texttt{choice} key. Its adapter also accepts a bare supplied label or a one-field JSON or Python-literal dictionary using a declared alias (\texttt{image}, \texttt{selected}, \texttt{selected\_image}, or \texttt{candidate}), with a value consisting solely of a supplied label, optionally preceded by ``Candidate''. Every alias, bare label, or non-JSON literal is marked as a recovery. Multiple fields, multiple labels, absent candidates, explanatory prose, and ambiguous values are rejected. The adapter never invents a ranking or consults human scores. Its output instruction replaces the ranking request with a choice-only contract while retaining the evaluation criteria. Raw output is retained for every attempt. Exceptions receive one identical retry; unresolved parse failures produce abstentions. Unique request keys combine prompt identity and permutation index, and resume checks require unchanged instruction and manifest hashes.

All public source revisions, model cards, candidate identities, frozen orders, environment versions, and derived tables are saved. Exact source-image bytes are retained locally; aspect-preserving resizing occurs only at inference. The pipeline accesses only public sources. Floating-point greedy decoding is deterministic within the recorded runtime configuration but need not yield identical choices across hardware or library implementations.
\end{document}